%% file: 0-MAIN.tex
\RequirePackage{fix-cm}
\documentclass[smallextended, final]{svjour3}       
\journalname{Data Mining and Knowledge Discovery}
\smartqed  
\usepackage[linesnumbered,ruled,vlined]{algorithm2e}
\usepackage{subfiles}
\usepackage{graphicx}
\usepackage{xcolor}
\usepackage{natbib}
\usepackage{comment}
\usepackage[backref=page]{hyperref}
\usepackage{booktabs}
\usepackage{amsfonts}
\usepackage{amsmath}
\usepackage{float}
\usepackage{svg}
\usepackage{makecell}
\usepackage{adjustbox}
\newcommand*\rot{\rotatebox{90}}
\usepackage{pdfpages}
\usepackage{mathtools,xparse}
\usepackage{subfig}
\usepackage{lscape}

\DeclarePairedDelimiter{\abs}{\lvert}{\rvert}
\DeclarePairedDelimiter{\norm}{\lVert}{\rVert}

\newcommand{\overbar}[1]{\mkern 1.5mu\overline{\mkern-1.5mu#1\mkern-1.5mu}\mkern 1.5mu}

\begin{document}
\title{Scalable Classifier-Agnostic Channel Selection for Multivariate Time Series Classification
}

\titlerunning{Scalable Classifier-Agnostic Channel Selection for MTSC}        

\author{Bhaskar Dhariyal    \and
        Thach Le Nguyen \and 
        Georgiana Ifrim
}


\institute{Bhaskar Dhariyal, Thach Le Nguyen and Georgiana Ifrim are with the Insight Centre for Data Analytics, School of Computer Science, University College Dublin, Dublin, Ireland;
        \{bhaskar.dhariyal,thach.lenguyen,georgiana.ifrim\}@insight-centre.org 
}
\date{Received: date / Accepted: date}

\maketitle


\abstract{Accuracy is a key focus of current work in time series classification. However, speed and data reduction are equally important in many applications, especially when the data scale and storage requirements rapidly increase. Current multivariate time series classification (MTSC) algorithms need hundreds of compute hours to complete training and prediction. This is due to the nature of multivariate time series data which grows with the number of time series, their length and the number of channels. In many applications, not all the channels are useful for the classification task, hence we require methods that can efficiently select useful channels and thus save computational resources. We propose and evaluate two methods for channel selection. Our techniques work by representing each class by a prototype time series and performing channel selection based on the prototype distance between classes. The main hypothesis is that useful channels enable better separation between classes; hence, channels with a longer distance between class prototypes are more useful. On the UEA MTSC benchmark, we show that these techniques achieve significant data reduction and classifier speedup for similar levels of classification accuracy. Channel selection is applied as a pre-processing step before training state-of-the-art MTSC algorithms and saves about 70\% of computation time and data storage with preserved accuracy. Furthermore, our methods enable efficient classifiers, such as ROCKET, to achieve better accuracy than using no selection or greedy forward channel selection. To further study the impact of our techniques, we present experiments on classifying synthetic multivariate time series datasets with more than 100 channels, as well as a real-world case study on a dataset with 50 channels. In both cases, our channel selection methods result in significant data reduction with preserved or improved accuracy.}


\subfile{1-introduction}
\subfile{2-Lit}
\subfile{3-method}

\subfile{4-expt}

\subfile{5-synth}
\subfile{6-usecase}

\subfile{8-limit}
\subfile{9-conclusion}

\begin{acknowledgements}
This publication has emanated from research supported in part by a grant from Science Foundation Ireland through the VistaMilk SFI Research Centre \\ 
(SFI/16/RC/3835) and the Insight Centre for Data Analytics (12/RC/2289 P2). For the purpose of Open Access, the author has applied a CC BY public copyright licence to any Author Accepted Manuscript version arising from this submission. We would like to thank the reviewers for their constructive feedback. We would like to thank all the researchers that have contributed open source code and datasets to the UEA MTSC Archive and especially, we want to thank the groups at UEA and UCR who continue to maintain and expand the archive.
\end{acknowledgements}

%
%

\bibliographystyle{spbasic}      
\bibliography{ref}   
\newpage
\subfile{appendix}

\end{document}

%% file: 1-introduction.tex
\section{Introduction}

Time-series data are ordered sequences of numeric values recorded over time, collected in vast quantities and used in many application domains. 
Low-cost sensors coupled with rapid advances in IoT devices and cloud infrastructure has fuelled time series data collection in agriculture \citep{riaboff2022predicting}, equipment monitoring \citep{kanawaday2017machine}, health \citep{portable-eeg}, sports\footnote{\url{https://www.outputsports.com/}} and fitness \citep{singhinterpretable}.

\begin{figure}[!h]
    \centering
    \includegraphics[width=\textwidth]{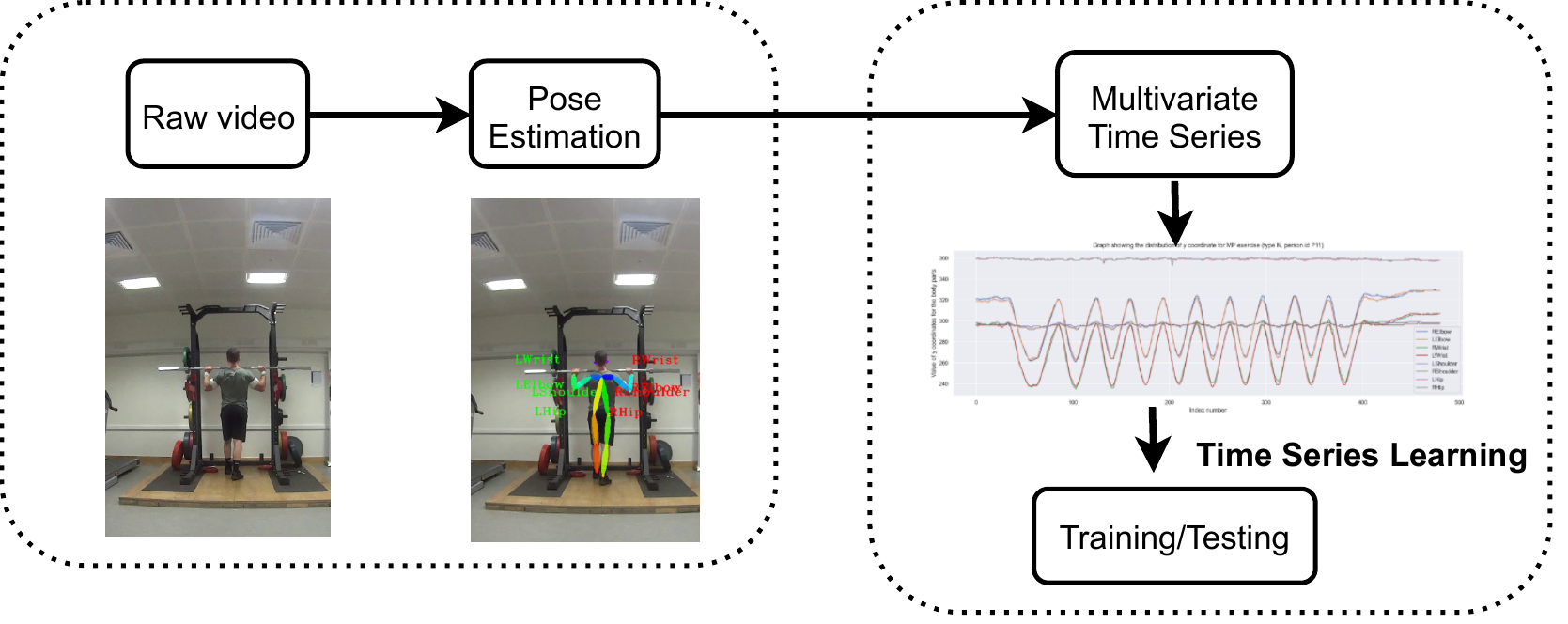}
    \caption{From video to multivariate time series using OpenPose (figure reproduced from \citep{singhinterpretable} with the permission of the authors).}
    \label{mp-tsc}
\end{figure}

The ease of collecting time-series data comes up with its own set of computational challenges. The scale of data collected increases  rapidly, and so do the computational resources required to store and analyse this data. For example, in equipment monitoring in the cloud, there can be thousands of samples collected per minute, millions of time series to be monitored, and the number of monitored services is growing rapidly \citep{DBLP:journals/pvldb/AdamsAABBBCCCJS20}.
Most time-series data are recorded in multivariate form where the temporal sequences come from multiple channels. For example, the Heartbeat dataset from the UEA MTSC benchmark has 61 channels. Each channel is the heartbeat's sound recorded at different locations in the body. In our real-world case study, the time series have 50 channels. In this setting, an exercise named Military Press (MP) is performed, where a person lifts a barbell above their head from shoulder level, and the tracked motion data is collected from different body parts. 
Each body part tracked along an XY-axis acts as a channel and the primary objective is to classify the multivariate temporal sequences into normal and aberrant subtypes (see Figure \ref{mp-tsc} for an illustrative example). For this problem, only a subset of the body parts (e.g., channels corresponding to the upper body) is useful for the classification task. In our experiments we show that by selecting a useful subset using our channel selection methods, we benefit from data reduction and also significantly gain in accuracy, as the noisy channels that can confuse the classifier are removed before training the classifier.

To evaluate the impact of channel selection, we work with recent 
multivariate time series classifiers, ROCKET \citep{dempster2019rocket}, Weasel-Muse \citep{muse18alltd} and MrSEQL-SAX \citep{le2019interpretable}. These approaches were shown to have state-of-the-art accuracy and can complete training and prediction on the full UEA MTSC benchmark in less than 7 days \citep{dhariyal20icdmw,ruiz2020great}. 
One important aspect of the classifiers mentioned above is that they use data from all the channels during training, and it is evident from our experiments that this is not the best strategy. The data from only a few channels could be more than sufficient to achieve the same or higher accuracy, thus reducing the time and memory required by the classifiers. For example, the DuckDuckGees dataset has 1,345 channels, but using only 30\% of those channels (as selected by our methods) actually improve the accuracy of the classifier while reducing the data significantly. Similar impacts of our methods can be observed in many of the other UEA MTSC datasets.

This work extends the preliminary study \citep{dhariyalfast21} on fast channel selection for multivariate time series classification. Although the primary focus of previous and current studies remains the same, i.e., improving classifier scalability while preserving accuracy, in this study we aim to achieve this along with maintaining robustness to noise. 
We achieve this goal by extending the previous channel selection methodology by studying different types of prototypes for representing classes, refining our channel selection and ranking methods and studying robustness to common data challenges, e.g., noise, minority classes and concept drift. 

We evaluate our methods on the popular UEA MTSC benchmark and on synthetic time series, which simulate different types of signals, e.g. gaussian, pseudo-periodic, autoregressive and diverse data generation processes, where the relevant points are static or are moving, to simulate concept drift. These simulated datasets help us to analyse the behaviour of channel selection across different domains. For example, EEG waveforms are generally pseudo-periodic~\citep{sainio1983eeg}, finance data usually follows auto-regressive patterns~\citep{chou2005forecasting}, while vibration mechanics involves a lot of Gaussian processes~\citep{avendano2017gaussian}.   
We also include a real-world case study on a human motion dataset collected as multivariate time series with 50 channels. We encounter challenges such as input noise due to the data collection process, minority classes, and relevant patterns that can shift for different time series due to the different motion characteristics of different people.

The primary objective of this study is to perform fast channel selection leading to data reduction without compromising on accuracy. 
We aim to achieve this by removing unnecessary channels, so that existing MTSC algorithms can perform classification efficiently and accurately.

\textbf{The main contributions of this work are:}
\begin{itemize}
 \item We present new channel selection techniques, which are fast, scalable, robust to different types of noise in the input data and classifier-agnostic.
\item We conduct extensive experiments on the UEA MTSC benchmark and report 75\% reduction in computation time and a data reduction of 60\% using our approaches, while preserving accuracy.
\item We demonstrate a significant improvement in the accuracy (more than 20\%) of the state-of-the-art classifier ROCKET on synthetic datasets with varying levels of  difficulty in identifying important channels. 
\item We show significant accuracy improvement for ROCKET (more than 5\%) on a real-world case study of our methods on a human motion MTSC dataset with 50 channels, recorded for classifying the Military Press strength and conditioning exercise.
\end{itemize}

%% file: 2-Lit.tex
\section{Related Work}
\label{sec:relwork}

In this section we give a brief overview of state-of-the-art MTSC methods, and discuss existing approaches for channel selection.

\subsection{Multivariate Time Series Classification} 

The recent empirical surveys \citep{ruiz2020great, dhariyal20icdmw} provide a detailed overview of progress in MTSC. This section describes a subset of the classifiers discussed in these surveys that were used in this work to evaluate the impact of channel selection methods.
We select these classifiers based on their accuracy on established benchmarks, availability of open source code and the computation time required to complete training and prediction on the UEA MTSC benchmark (less than 7 days).

\subsubsection{ROCKET} 

ROCKET \citep{dempster2019rocket} is one of the most accurate time-series classifiers, initially proposed  for univariate time series classification and later extended to MTSC. ROCKET relies on convolutional kernels to extract time-series features. The kernels in ROCKET are generated randomly; however, unlike in Convolutional Neural Networks, they are static throughout the classification process. The kernel properties like length, dilation, stride, bias and zero-padding are sampled randomly. The MTSC version of ROCKET also performs channel selection randomly, selecting a maximum of 12 channels for any MTSC dataset. Therefore, the runtime is not significantly affected by the number of channels, especially for datasets with more than 12 channels.

The kernels are a linear transformation of the input time series. A linear classifier, ridge regression, trains on the feature vector formed by global max pooling and the proportion of positive values  (PPV) features extracted from the convolution time series from every channel.
ROCKET has become very popular due to its high accuracy and speed, yet the impact of channel selection on this classifier in the MTSC task has not been examined extensively.

\subsubsection{MrSEQL-SAX}

The linear classifier MrSEQL \citep{LeNguyen2019} extracts symbolic features from the time series. The method transforms time-series data to multiple symbolic representations of different domains (e.g., SFA~\citep{sfa.2012} in the frequency domain and  SAX~\citep{lin-sax:dmkd07} in the time domain) and into different resolutions using various window sizes. The classifier extracts discriminative subsequences from the symbolic representations, and these subsequences are later combined to form a single feature vector used to train a classification model. The method was initially developed for univariate time series classification but later adapted to MTSC; the adapted version views each channel as an independent time series. 
The symbolic transformations are computationally expensive. Furthermore, the transformation over multiple windows iterating over full-time series incurs a high cost for the scalability of the classifier. Therefore, reducing the number of channels can significantly improve the classifier's scalability.  This study focuses on MrSEQL-SAX (MrSEQL using only SAX features) as it is more efficient than the variant combining SAX and SFA transformations. 

\subsubsection{WEASEL-MUSE} 
WEASEL-MUSE \citep{muse18alltd} is an extension of the WEASEL algorithm \citep{DBLP:conf/cikm/SchaferL17} to MTSC data. The classifier builds a bag-of-patterns (BOP) model using the SFA transform for every channel. It captures the patterns by rolling multiple windows of varying sizes on raw and derivative time series, and transforming those segments into unigram and bigram words. The classifier links these words to their respective channel and creates a histogram for each channel separately. To deal with the large number of features from every channel, the Chi-square feature selection method removes the irrelevant features. The selected features are concatenated into a single feature vector which is the input to a logistic regression algorithm.
Like MrSEQL, the WEASEL-MUSE classifier also iterates over the entire time series for every channel and performs the SFA transform for every window. This iteration and transformation increases the overall computation cost. Also, storing many unigrams and bigrams in memory is quite expensive.  Therefore, we expect that channel selection would positively impact this classifier.


\subsection{Channel Selection for Multivariate Time Series Classification}

Channel selection for multivariate time series is a recurring topic in the MTSC literature. However, the focus of most work has been on accuracy rather than scalability. 

The most recent work on channel selection \citep{kathirgamanathan2020feature} aims to identify the best subset of channels. The proposed method calculates a merit score based on correlation patterns of the outputs from the classifiers. The algorithm iterates through every possible subset to calculate the merit score, followed by a wrapper search on the subset with the top 5\% merit score. 1NN-DTW is employed as the wrapper to perform the classification. The computational bottleneck of this work is using DTW over every possible subset to identify the useful channels. 

Another notable study is CleVer \citep{yoon2005feature} where the author proposed three unsupervised feature subset selection techniques employing Common Principal Component Analysis (CPCA)\citep{krzanowski1979between} to measure the importance of each sensor. The authors build a correlation coefficient matrix among different channels for each MTS. The principal components of each coefficient matrix are calculated, all the principal components are aggregated together, and descriptive component principal components are calculated. The $l_2$-norm of the resulting vector generates the rank of each channel. 

\cite{hu2013icdm} proposed a framework for channel selection using a voting-based method. The two criteria used were distance-based classification and confidence-based classification. These methods were proposed for streaming data, which is outside the scope of the current study. 

A recent study \citep{han2020supervised} presents an algorithm for channel ranking and channel selection. 
The channel ranking algorithm assigns a relevance score for each channel. The relevance scores are constructed on a similarity graph among the channels. The authors find the largest eigenvector of the normalized adjacency matrix of the similarity graph, which reflects its cluster structure. Apart from channel ranking, the authors also propose channel subset selection. From the adjacency matrix above, the algorithm finds the linear combination of matrices that approximates the labels' similarity matrix and uses the minimum number of redundant channels. Although the proposed channel ranking and selection approach performs well regarding the accuracy, it is slow. The authors mention that they face computation issues while building the graph representation for each feature.

The works mentioned above perform channel selection at the cost of computation time, sometimes leading to significantly increased time as compared to the default classifiers without channel selection. We note that none of the methods described above have open-source code available to perform comparative analysis. In order to compare our methods to a strong baseline, we adapt the feature selection methods from scikit-klearn\footnote{\url{https://scikit-learn.org/stable/modules/generated/sklearn.feature_selection.SequentialFeatureSelector.html}} to work with multivariate time series, by considering each channel as a single feature. We also investigate another baseline that uses the $l_2$-norm of each channel as a criteria for ranking and selecting channels.
We provide all the data and code for reproducing our  results\footnote{\url{https://github.com/mlgig/ChannelSelectionMTSC}}.

%% file: 3-method.tex
\section{Methods}
\label{sec:meth}

In this section we present our methodology for channel selection and discuss the key components by means of an example from our case study.

\begin{figure}
    \centering
    \includegraphics[width=\textwidth]{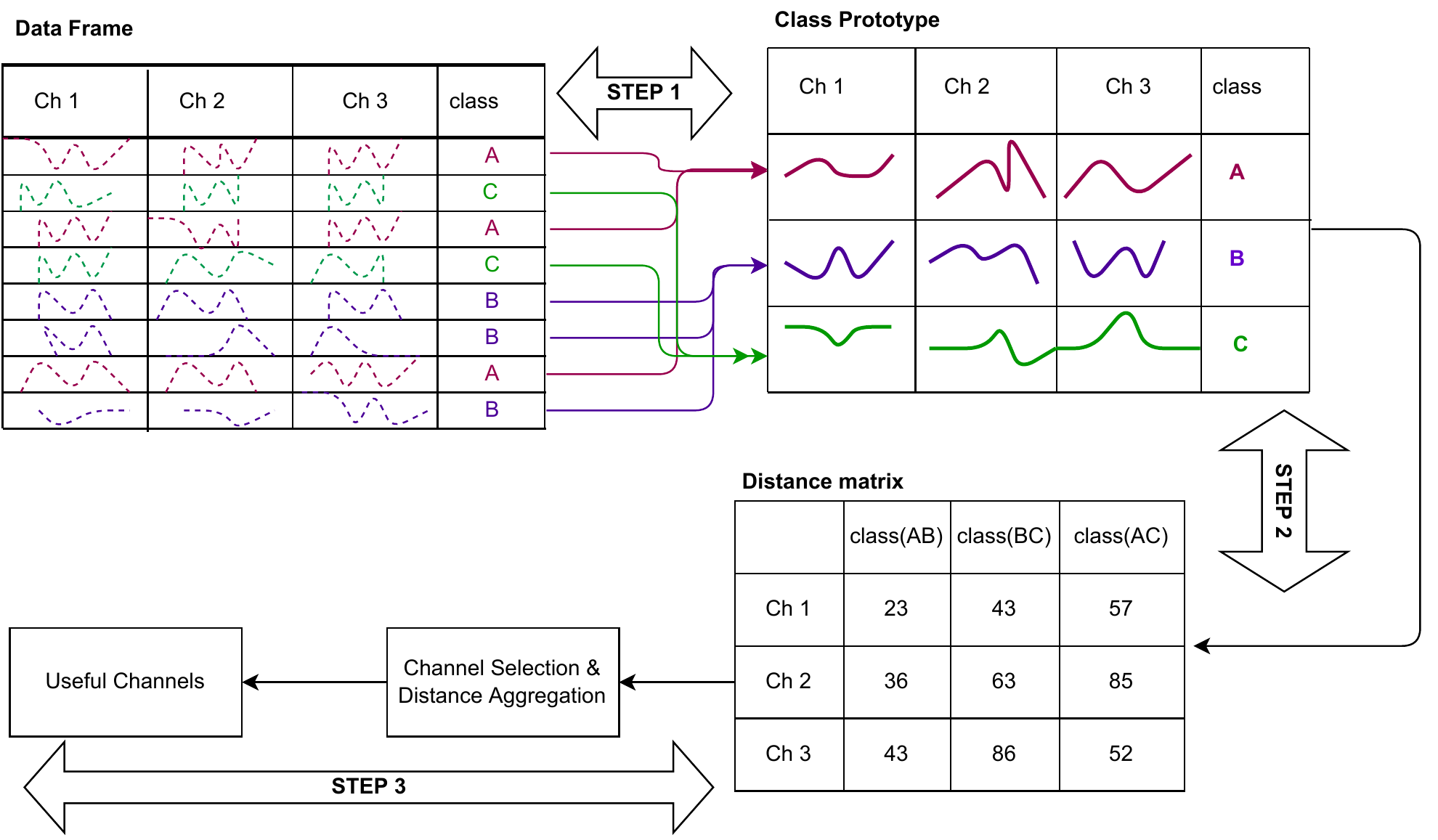}
    \caption{Channel Selection Pipeline: 1. Compute class prototypes for each class and each channel. 2. For each channel, compute distance between prototype pairs. 3. Select the channel subset with the highest distance between prototypes. Hypothesis: Channels with higher distance are more useful for separating the classes in the subsequent classification task.}
    \label{fig:pipeline}
\end{figure}

The proposed channel selection techniques depend on three components: \textit{class prototype}, \textit{distance matrix} and \textit{distance aggregation} (Figure~\ref{fig:pipeline}). Class prototypes are basically representations of each class in the dataset. The class prototype is also a multivariate time series. The key idea of this approach is that by comparing class prototypes for each channel, we can get an indication of which channel is more useful. The channels that increase the distance between class prototypes are more likely to be useful for the classification task, since they add in class discrimination.
In the next step we compute a distance matrix between prototypes (pairwise distances between class prototypes) and use distance aggregation to rank and select channels. Next we discuss different choices of class prototypes and how to employ the distance matrix to select useful channels.

\subsection{Class Prototypes}
\label{subsec:cp}
A representative prototype of a class is necessary for the proposed techniques. In our approach, it is also important that the class prototypes are inexpensive to compute. Extending our previous work \citep{dhariyalfast21} on channel selection, here we investigate other choices for the class prototypes, with a specific focus on robustness to noise. We focus on three prototypes: mean, median and MAD (data trimmed using median absolute deviation \citep{leys2013detecting}). Next, we formally define the class prototypes.

Let $X \in R^{ n\times d \times l}$ be an MTS dataset and $y$ the labels of the time series in the dataset; $n$ represents the number of time series in the dataset, $d$ represents the number of channels in the multivariate time series, and $l$ is the length of the time series. 

\textbf{Mean Prototype:} Let $X_A = \{s \in X \mid y(s) = A\} $ be the subset of $X$ that contains only samples from class A. The mean prototype of class A is computed as the \textit{mean} of all the time series in that class, i.e., the centroid:
\begin{equation}
    C_A[i,j] = \frac{\sum_{k=1}^{k=m} X_A[k,i,j]}{m}
\end{equation}
where $m$ is the number of samples in class $A$. The multi-channel mean prototype $C_A$ is a $d \times l$ matrix in which each row $C_{A,i}$ is the centroid of class $A$ for channel $i$.

\textbf{Median Prototype:} Mean estimation is known to be susceptible to outliers. Median is an alternative to remedy this issue. For class A, the median prototype is computed as:

\begin{equation}
    M_A[i,j]=calculate\_median({X_A[k,i,j]}^{k=m}_{k=1})
    \label{eqn:median}
\end{equation}
where $m$ is the number of samples in class $A$. The multi-channel median prototype $M_A$ is a $d \times l$ matrix in which each row $M_{A,i}$ is the median-center of class $A$ for channel $i$.
From hereon, we refer to the median prototype as the median for simplicity.

\textbf{MAD prototype:} Another method to handle outliers is clamping, i.e., replacing outliers with predetermined threshold values. We propose the MAD prototype for multivariate time series, which clamps the outliers using Median Absolute Deviation (\cite{leys2013detecting}). The MAD of class A is computed as follows:

\begin{equation}
   mad_{A}[i,j] = median(\vert X_A[k,i,j] - M_A[i,j]\vert^{k=m}_{k=1}) 
\end{equation}

The upper and lower thresholds for clamping are then defined as

\begin{equation*}
    \textit{upper\_limit} = M_{A} + 0.5*mad_{A}
\end{equation*}
\begin{equation*}
    \textit{lower\_limit} =  M_{A} - 0.5*mad_{A}
\end{equation*}




For a channel $i$, the MAD prototype for class $A$ is calculated as the mean of the clamped dataset $\overbar{X}$ (values outside of the $[\textit{lower\_limit}, \textit{upper\_limit}] $ range are replaced with the respective boundary values).

\begin{equation}
    \text{MAD}_{A}[i,j] = \frac{\sum_{k=1}^{k=m} \overbar{X}_A[k,i,j]}{m}
\end{equation}

 The multi-channel MAD prototype $MAD_A$ is a $d \times l$ matrix in which each row $MAD_{A,i}$ is the (clamped) centroid of class $A$ for channel $i$. For the purpose of simplicity we refer to this prototype as MAD.

\begin{figure}
\centering
\subfloat[]{\includegraphics[width=0.8\textwidth]{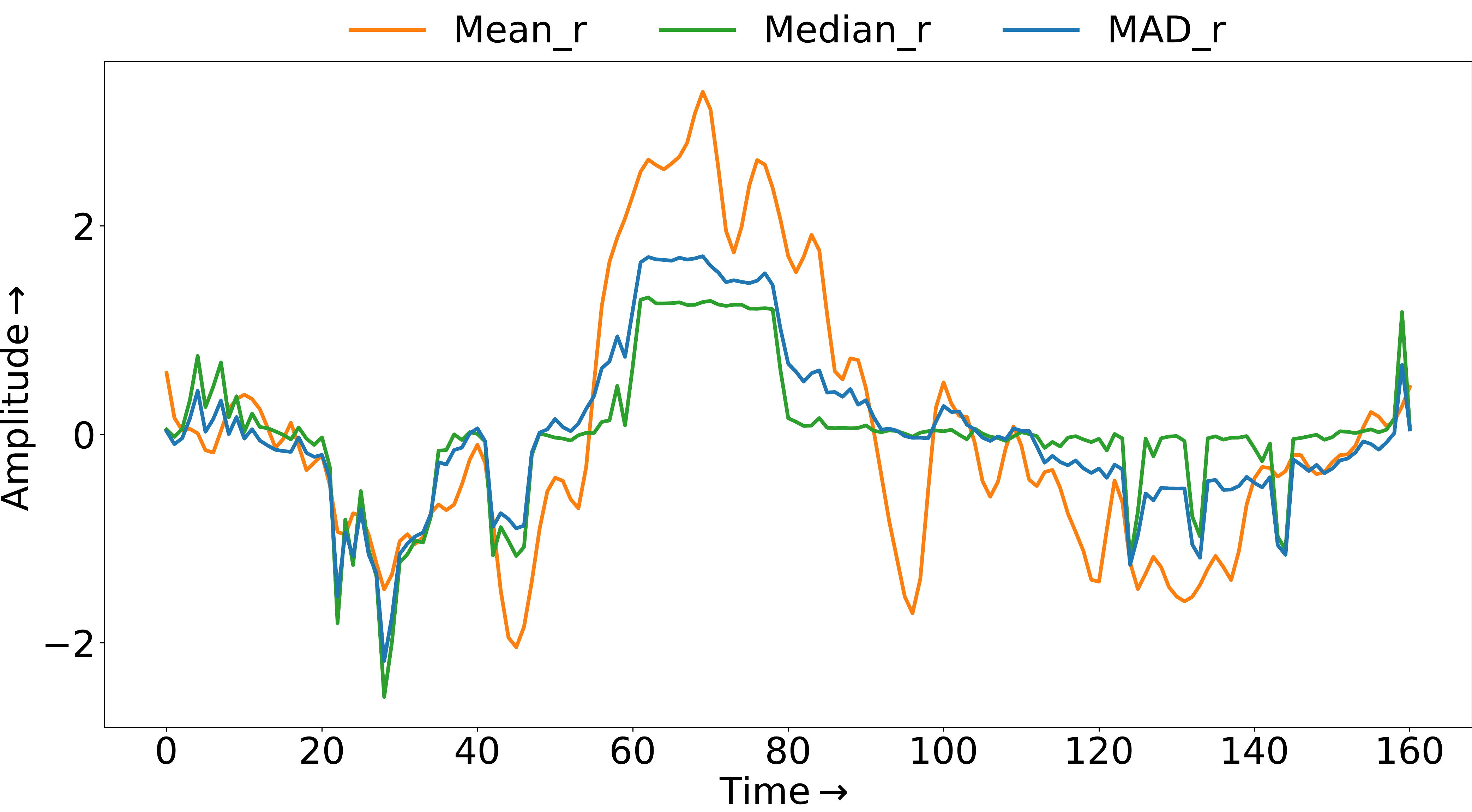}}
   

\subfloat[]{\includegraphics[width=0.8\textwidth]{figures/LWrist_X_r.pdf}}

\caption[]{Class prototypes for the classes normal (n) and reduced (r) for the left wrist channel along the X-axis (details of this dataset are given in the case study Section \ref{sec:case_study}).}
\label{fig:prot_eg}
\end{figure}

Figure \ref{fig:prot_eg} illustrates the prototypes for two classes in the Military Press dataset. We notice that the mean prototype is more volatile than the other two, which can be explained by the outliers in the data. For this task, the time series data is obtained using body pose estimation libraries, and when the body is not fully visible or the estimation is not accurate, there can be outliers in the data.
On the other hand, median and MAD prototypes follow each other closely. However, the MAD prototype seems to be more reflective of changes in the data.  

\subsection{Distance Matrix} 

We compare the class prototypes by measuring their Euclidean distance for each channel and for each pair of classes. The result is a distance matrix, which is a two-dimensional array of channels and class pairs. In Figure \ref{fig:pipeline}, the distance matrix tells us that the (Euclidean) distances between class A and class B prototypes in channel 1, 2, and 3 are 23, 36, and 43, respectively. This implies that channel 3 might be more useful than channel 1 and 2 in separating class A and class B. Table \ref{tab:dm_mad} illustrates a real distance matrix for the Military Press dataset. It is important to note that we also tested DTW as an alternative distance measure, which is usually preferred in time series analysis. Nevertheless, DTW is more expensive, hyperparameter sensitive, and its benefit when compared to Euclidean distance was unnoticeable in our experiments (see Appendix for details results in Tables \ref{tab:dtweu_acc} and \ref{tab:dtweu_time}). 



 

\subsection{Distance Aggregation and Channel Selection}

\subsubsection{The ECS and ECP Channel Selection Algorithms}

The preliminary study \cite{dhariyalfast21} proposed three techniques to perform channel selection using a simple class centroid as prototype, namely, KMeans, ECS (Elbow Class Sum), and ECP (Elbow Class Pairwise). Since KMeans performed poorly in comparison with ECS and ECP, we did not include it in this study. Here, we present the pseudo-code for our two algorithms ECS and ECP and present new extensions of our methodology in order to be more robust to data noise. In particular, we investigate new class prototypes, different ways to compute and aggregate the distance matrix and expand the channel selection and ranking strategy.

\begin{algorithm}[h]
\SetAlgoLined
\KwIn{Training dataset: X,y}
\KwOut{Selected channels, ranking of channels}
 Initialization\;
 For each channel in X and each class, compute class prototype\;
 Compute distance matrix\;
Sum the distance matrix across rows\;
Sort channels by sum in decreasing order\;
Find the elbow on the sorted channels\;
Selected channels = channels with sum $>$ elbow point \;
Rank channels by total distance summed over class pairs\;
Return Selected channels and their ranks\;
 \caption{ECS Channel Selection for an MTSC dataset.}
 \label{algo_ecs}
\end{algorithm}

\begin{algorithm}[h]
\SetAlgoLined
\KwIn{Training dataset: X,y}
\KwOut{Selected channels, ranking of channels}
 Initialization\;
 For each channel in X and each class, compute class prototype\;
 Compute distance matrix\;
For each column in the distance matrix\;
\hspace{4mm} Rank the channels in decreasing order of the distance in the corresponding column of the distance matrix\;
\hspace{4mm} Find the elbow on the ranked channels\;
\hspace{4mm} Selected channels = Selected channels $\cup$ channels with distance $>$ elbow point\;
Rank channels by total distance summed over class pairs\;
Return Selected channels and ranks\;
 \caption{ECP Channel Selection for an MTSC dataset.}
 \label{algo_ecp}
\end{algorithm}

Algorithms \ref{algo_ecs} and \ref{algo_ecp} describe the distance aggregation and channel selection steps for the ECS and ECP methods respectively. ECS evaluates each channel by the total sum of pairwise distances between class prototypes for that channel. However, for some tasks, there are classes that are more distinguishable than the others, hence the distance-sum can be dominated by the distances between these classes. As a result, channels that are specifically more useful for the less distinguishable classes might be ignored. To address this issue, ECP iterates through every class pair and selects the best channels for each pair. The final set of selected channels is a union over the selection for each pair.
In this way, ECP ensures that the channels that might be ignored in ECS, also have a good chance to be selected.

In addition, we have evaluated a simpler baseline technique to rank channels using their \textit{l2-norm}. The basic idea of using the $l_2$-norm is that channels with higher magnitude might be more useful for discriminating between classes 

    
\subsubsection{Elbow Cut on Ranked Channels}

\begin{figure}[H]
    \centering
    \includegraphics[width=\textwidth]{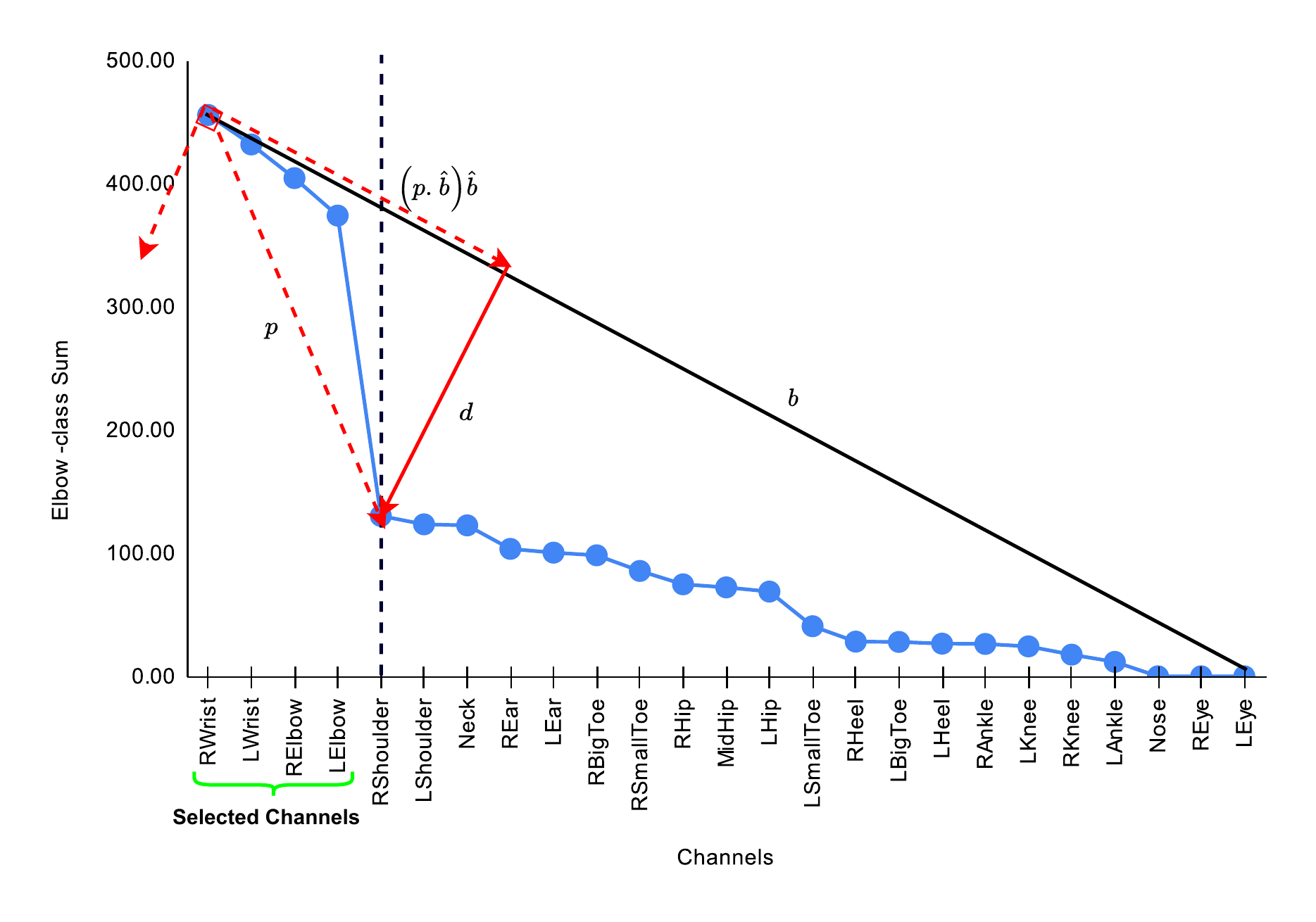}
    \caption{Elbow-cut example for channel selection on the Military Press dataset using the \textit{MAD} class prototype.}
    \label{fig:elbowcut}
\end{figure}

Both the ECS and ECP algorithms employ an heuristic for automatically selecting a set of channels from a given ranked list of channels. This heuristic is commonly known as the elbow cut.
The elbow cut is used to select channels from the distance matrix (ECP) or the distance sums (ECS). It identifies an inflection point on a ranked list of distances (Figure \ref{fig:elbowcut}) and returns the channels based on this point (commonly known as the elbow point).


The elbow point (see Figure \ref{fig:elbowcut}) is the point at the highest distance \textit{d} from the line \textit{b} joining the initial and ending point in a ranked list. The distance $d$ to any point on the distance curve is calculated by using a vector projection of $p$ on the line $b$ as shown in Equation \ref{eqn:dis}.

\begin{equation}
d= \abs{p-(p.\hat{b})\hat{b}}    
\label{eqn:dis}
\end{equation}

 where $\hat{b}$ ($=\frac{b}{\norm{b}}$) is the unit vector in the direction of line $b$.  The $p.\hat{b}$ is the scalar projection of $p$ onto $\hat{b}$ (\cite{perwass2009geometric}) and Equation \ref{eqn:dis} is obtained by the triangle formula. Thus, the elbow-point is 
 \begin{equation}
  \text{elbow} = argmax(d)   
 \end{equation}
The channels before the $elbow$ are selected as useful channels (highlighted in green) for classification, and the smaller dataset with this subset of channels is used for the classification step. It is clear from Figure \ref{fig:elbowcut} that the elbow point can be relaxed, thus allowing a trade-off between data storage and the accuracy of classification.

\subsubsection{Channel Ranking} Although the proposed channel selection reduces the system's computation requirements, it is not always possible to allocate a clear meaning to selected channels without domain experts. This work presents a method to rank channels according to their discriminative power (where distance between classes is used as a proxy for discrimination). ECS directly works on the sorted distances; therefore, it is easier to rank channels according to the distances. In the case of ECP, the channels selected by ECP are finally ranked based on the sum of distances across classes, similar to the first step in ECS.

\subsection{Computational Complexity}
\label{sec:comp_comp}

As mentioned in Section \ref{subsec:cp} the dataset is defined with $n$, $d$ and $l$ as parameters for the number of time series, number of channels, and length of time series. Let $c$ be the number of the classes. To analyze the time complexity for our proposed algorithms, we break the techniques into three steps, as illustrated in Figure \ref{fig:pipeline}, namely: 1. computation of class prototypes, 2. distance matrix formulation, and 3. channel selection.

The time complexity to create the class prototypes is $ \sim\mathcal{O}(n \times l \times d)$. The distance matrix stores the distance for each channel for every class pair; therefore, the complexity of computing the distances for each class combination drives the time complexity. Theoretically the time complexity to create a distance matrix with $d$ channels is $ \sim\mathcal{O}(c^2 \times l \times d)$.
In the final step, the elbow cut takes $ \sim\mathcal{O}(d \times log(d))$ to sort and perform elbow cut in ECS and $ \sim\mathcal{O}(c^2 \times d \times log(d))$ in ECP. 

The total time complexity for channel selection is then the sum of the three steps: computer prototypes + computer class pairwise distances for each channel + sort distances and compute elbow point, which is $\sim\mathcal{O}(n \times l \times d) +  \mathcal{O}(c^2 \times l \times d) + \mathcal{O}(c^2 \times d \times log(d))$. In practice, $d$ and $c$ are typically small numbers in comparison to $n$ and $l$. It is very rare for a multivariate time series dataset to have more than tens of classes or hundreds of channels. Therefore it is acceptable to assume that $log(d) << l$ and $c^2 < n$. As a result, the final complexity is  $\sim \mathcal{O}(n \times l \times d)$. Table \ref{tab:timespace} show the empirical time and memory results for ECS and ECP and reflects the theoretical complexity derived here.

Regarding space complexity, the only additional information stored is the distance matrix, which grows $\sim\mathcal{O}(c^2 \times d)$. This is insignificant given that the values of $d$ and $c$ are typically small.

%% file: 4-expt.tex
\section{Evaluation on Benchmark Datasets}

The approaches designed in this study are evaluated using state-of-the-art multivariate time series classification algorithms from the popular Python library sktime (\cite{loning2019sktime}). It should be noted that the algorithms are not tuned, but are used with the default parameters recommended in the original papers. However, our objective in designing experiments is to understand the relative gain or loss in computational aspects of MTSC algorithms using the proposed channel selection techniques instead of benchmarking MTSC algorithms. 
All the experiments were run on a Intel(R) Xeon(R) Gold 6140 CPU server with 256GB RAM. We release all our data and code in our Github repository\footnote{\url{https://github.com/mlgig/ChannelSelectionMTSC}}.

\subsection{Datasets}
\label{subsec:data}
The UEA/UCR Time Series Classification archive \citep{bagnall16bakeoff} is a collection of univariate and multivariate time series data. The repository contains 30 multivariate datasets from various application domains, e.g. ECG, motion classification,  spectra classification. These heterogeneous datasets vary regarding the number of channels (from 2 to 1,345), the number of time series (12 to 30,000) and time series length (8 to 17,894). Here we work with the subset of 26 datasets with equal-length time series. The datasets have pre-defined splits into train and test sets and we use these splits for our evaluation. We call this archive the UEA MTSC benchmark.

\subsection{State-of-the-art MTSC Algorithms}
\begin{figure}[H]
    \centering
    \includegraphics[width=\textwidth]{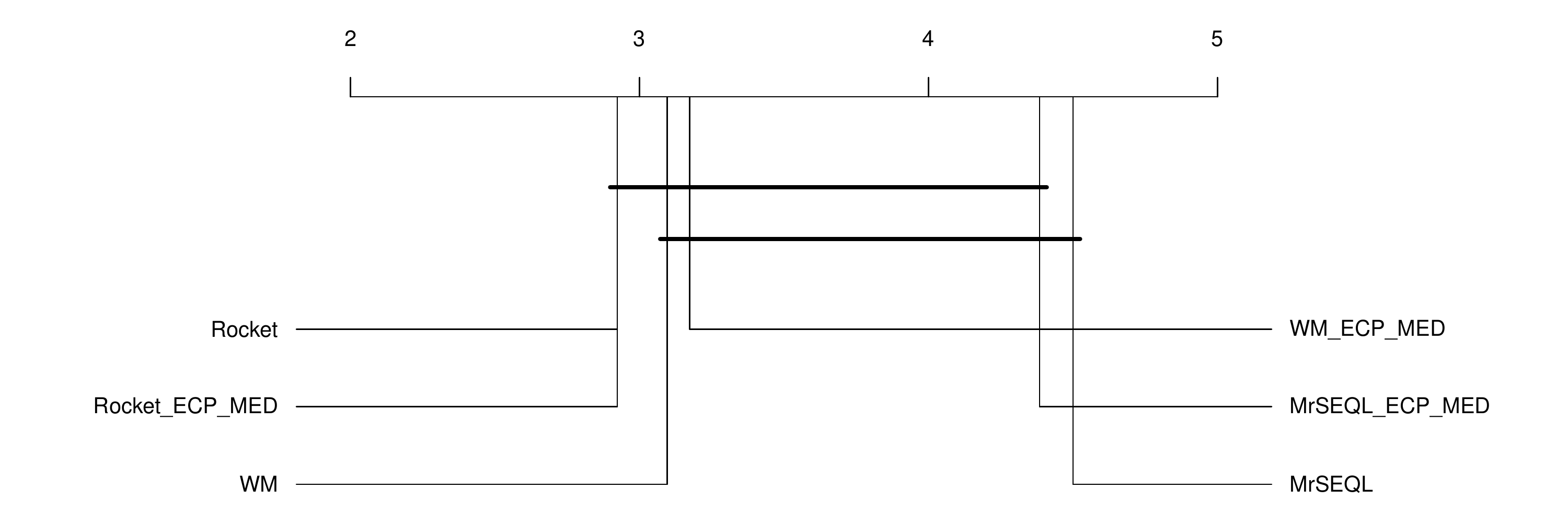}
    \caption{Performance of MTSC classifiers with and without channel selection.}
    \label{fig:cs_best}
\end{figure}

Before analyzing the performance of channel selection, we benchmark existing state-of-the-art algorithms on our system. Table \ref{tab:sota_perf} shows the performance of classifiers on the 26 datasets. For ease of comparison we first show the average accuracy over the 26 datasets, as well as the total runtime to complete training and prediction over all datasets. In order to test for the statistical significance of these results, we follow the recommendations \citep{Demsar:2006,garcia-extension,JMLR:v17:benavoli16a}. The accuracy gain is evaluated using a Wilcoxon signed-rank test with Holm correction and visualised with the critical difference (CD) diagram. The CD shows the ranking of  methods with respect to their average accuracy rank computed across multiple datasets. 
Methods that do not have a  statistically significant difference in rank, are connected with a thick horizontal line. 
For computing the CD we use the R library \textit{scmamp}\footnote{\url{https://github.com/b0rxa/scmamp}} \citep{scmamp}. 
ROCKET is the best method for scalability and accuracy, closely followed by WeaselMuse and MrSEQL-SAX. However, WeaselMuse and MrSEQL-SAX are slow in comparison to ROCKET. For comparison, in Table \ref{tab:sota_perfcs} we show the performance of these classifiers after channel selection with our methods. We also see in Figure \ref{fig:cs_best} that there is no significant difference between classifier accuracy before and after channel selection, i.e., with fewer channels and less computation time. We describe next a comparison of our methods to a strong greedy feature selection baseline, as well as a sensitivity analysis for the main components in our methods.

\begin{table}[!h]
    \centering
    \begin{tabular}{c|c|c}
    \hline
         Classifiers &  Avg. Accuracy & Time (minutes) \\
         \hline
         Rocket & 71.58 & 36.93 \\
         WeaselMuse & 70.28 & 3774.47\\
         MrSEQL-SAX &  66.99 &  7314.05\\
         \hline\hline
         Total Memory (data size) & 1.2 GB\\
         \hline
    \end{tabular}
    \caption{Performance of state-of-the-art multivariate time series classifiers on 26 UEA datasets before channel selection.}
    \label{tab:sota_perf}
\end{table}

 \begin{table}[!h]
     \centering
     \begin{tabular}{c|c|c}
     \hline
          Classifiers &  Accuracy & Time (minutes) \\
          \hline
          Rocket &  73.01 & 29.26  \\
          WeaselMuse & 71.26 & 1104.10 \\
          MrSEQL-SAX &  68.30 & 2818.70 \\
          \hline \hline
          Total Memory (data size) & 0.38 GB\\
          \hline
     \end{tabular}
     \caption{Performance of state-of-the-art multivariate time series classifiers on 26 UEA datasets after channel selection. The best combination of prototype and channel selection approach is used for computing these values. For more details on best performance for each dataset please see Table \ref{tab:acc_aftercs}.}
     \label{tab:sota_perfcs}
 \end{table}


\section{Baseline Comparison: Forward Channel Selection  }

Filter and  Wrapper methods \citep{john1994irrelevant} are the methods of choice for performing feature selection in tabular data \citep{DBLP:journals/jmlr/GuyonE03}. The difference between the two methods is that the wrapper methods are exposed to classifier performance while the filter methods are agnostic of any classification evaluation metrics.
A wrapper method was proposed in \citep{kathirgamanathan2020feature} but no code was made available to facilitate comparison, and it is apparent from that work that the method would be computationally expensive.
We modify the forward feature selection in sckit-learn to a forward channel selection (FCS) method, where a channel of multivariate time series replaces a feature. There are two types of wrapper methods, namely, forward selection and backward elimination. This study uses tweaked forward selection as a baseline method. We deliberately leave the backward elimination method out as it is slow and requires a lot more memory when compared with forward selection. In default forward selection, the method greedily evaluates channels using a classifier and cross-validation, and provides a ranked list as output. The default setting  selects the top 50\% channels. As an alternative, domain knowledge can be used to decide the number of channels to be selected. To make the baseline stronger (than the default 50\% selection), we tweak this method to select channels starting from the best channel and adding channels until there is no more gain in accuracy (the accuracy tends to increase when starting from the best channel and adding new channels, and then tends to decrease once less informative channels are added). This is similar in nature to using the elbow cut to cut the ranked list of channels.      
\begin{table}[!h]
    \centering
    \begin{tabular}{c|c|c}
    \hline
         Classifier &  Avg. Accuracy & Time (minutes) \\
         \hline
          ROCKET & 71.58 & 36.93 \\
        FCS-ROCKET & 70.96 & 3685.13 \\
         \hline
    \end{tabular}
    \caption{Forward Channel Selection using ROCKET as estimator on 26 datasets.}
    \label{tab:baseline}
\end{table}

Table \ref{tab:baseline} shows the performance of the forward channel selection with ROCKET. We find enough evidence of the performance of FCS as it clearly shows that the default classifier, ROCKET, is much faster and more accurate than the baseline, FCS-ROCKET. Therefore it contradicts the primary argument of this work, which is enabling the existing classifier to scale better.  It is not feasible to perform forward channel selection for other MTSC methods for computational reasons.

\subsection{Proposed Channel Selection Methods}

In this section we focus on analysing the impact of different class prototypes and channel selection algorithms on the accuracy and runtime of the MTS classifiers.
Table \ref{tab:sotacs} shows the performance of the Median class prototype along with the Mean class prototype. The column $\Delta$Acc shows the difference in accuracy between the default classifier and the one with channel selection. The \%Time column shows the amount of time saved due to channel selection.
The Median class prototype, and ECP perform better than the Mean class prototype. The ECP-Median channel selection achieves better performance than the default version of ROCKET and MrSEQL-SAX, additionally, saving computation time and memory.     

\begin{table}[!h]
    \centering
    \begin{tabular}{c|cc|cc}
    \hline
    \multicolumn{1}{c|}{Prototype$\rightarrow$}  & \multicolumn{2}{c}{Mean} &  \multicolumn{2}{c}{Median}\\
    \hline
         \multicolumn{1}{c|}{Channel Selection$\rightarrow$}  &  \multicolumn{1}{c}{ECS} & \multicolumn{1}{c|}{ECP} & \multicolumn{1}{c}{ECS} & \multicolumn{1}{c}{ECP} \\
    \hline
        Classifier$\downarrow$            & $\Delta$Acc $\vert$ \%Time & $\Delta$Acc $\vert$ \%Time & $\Delta$Acc $\vert$ \%Time & $\Delta$Acc $\vert$ \%Time \\ 
    \hline
    Rocket & \textcolor{red}{-4.41} $\vert$ 36.74  &\textcolor{blue} {0.00} $\vert$ 14.19 &\textcolor{red}{-5.02} $\vert$ 28.51 & \textcolor{blue}{+0.11} $\vert$ 18.30 \\
    WeaselMuse & \textcolor{red}{-3.80} $\vert$ 76.90 & \textcolor{red}{-1.57} $\vert$ 67.97  &\textcolor{red}{-6.18} $\vert$ 77.64 & \textcolor{red}{-1.32} $\vert$ 68.02\\
    MrSEQL-SAX & \textcolor{red}{-3.85} $\vert$ 71.74 &\textcolor{blue}{+0.45} $\vert$ 59.65 &\textcolor{red}{-3.67} $\vert$ 71.05 &\textcolor{blue}{+0.36} $\vert$ 59.66 \\
    \hline

    \end{tabular}
    \caption{Performance of channel selection on 26 equal-length UEA datasets.}
    \label{tab:sotacs}
\end{table}

\begin{table}[!h]
    \centering
    \begin{tabular}{c|c|c}
    \hline
     \multicolumn{1}{c|}{Prototype$\rightarrow$}  & \multicolumn{2}{c}{MAD}\\
      \hline
    \multicolumn{1}{c|}{Channel Selection$\rightarrow$}  &  \multicolumn{1}{c|}{ECS} & \multicolumn{1}{c}{ECP}\\
    \hline
       Classifier$\downarrow$ & $\Delta$Acc $\vert$ $\%$Time & $\Delta$Acc $\vert$ $\%$Time \\ 
        \hline
         ROCKET &\textcolor{red}{-4.53} $\vert$ 27.59 & \textcolor{blue}{+0.28} $\vert$ 17.67\\
        WeaselMuse & \textcolor{red}{-5.77} $\vert$ 83.19 & \textcolor{red}{-1.34} $\vert$ 74.39 \\
        MrSEQL-SAX & \textcolor{red}{-3.31} $\vert$  83.39 & \textcolor{blue}{+0.38} $\vert$ 77.11 \\
        \hline
    \end{tabular}
    \caption{Performance of channel selection with the MAD prototype on the 26 UEA datasets. ECP-MAD outperforms the other selection strategies.}
    \label{tab:madres}
\end{table}

Although ECP-Median improved the performance, both the class prototypes have their virtues, as described in Section \ref{sec:meth}. The MAD centroid combines both the Mean class prototype and Median class prototype properties. Table \ref{tab:madres} shows the results when using MAD as the class prototype. The computation time also reduces significantly in the case of Weasel-Muse and MrSEQL-SAX.

To further understand the impact of different selection strategies, we have tested another approach for ranking and selecting channels.
We first calculate the \textit{l2-norm} of class prototypes for each class instead of the euclidean distance between a class pair. Table \ref{tab:l2} illustrates the performance of the \textit{l2-norm} on two class-prototypes namely, Mean and Median. Here we examine if channel selection can be performed directly based on the channels' magnitude, rather than the distance between class prototypes.
Table \ref{tab:l2} illustrates the performance of using the \textit{l2-norm} on two-class prototypes namely, Mean and Median.

\begin{table}[!h]
    \centering
    \begin{tabular}{c|c|c|c|c}
    \hline
    \multicolumn{1}{c|}{Prototype$\rightarrow$}  & \multicolumn{2}{c}{Mean} &  \multicolumn{2}{c}{Median}\\
    \hline
         Channel Selection$\rightarrow$ &  ECS & ECP & ECS & ECP \\
        Classifier$\downarrow$            & $\Delta$Acc $\vert$ $\%$Time & $\Delta$Acc $\vert$ $\%$Time & $\Delta$Acc $\vert$ $\%$Time & $\Delta$Acc $\vert$ $\%$Time \\
        \hline
        ROCKET &\textcolor{red}{-6.03} $\vert$ 40.49 &\textcolor{red}{-0.21} $\vert$ 19.36 & \textcolor{red}{-4.58} $\vert$ 30.10 & \textcolor{blue}{+0.30} $\vert$ 21.30\\
        WeaselMuse &\textcolor{red}{-4.80} $\vert$ 80.31 &\textcolor{red}{-0.66} $\vert$ 69.07 & \textcolor{red}{-6.07} $\vert$ 78.06 & \textcolor{red}{-2.14} $\vert$ 72.04\\
        MrSEQL-SAX & \textcolor{red}{-5.14} $\vert$ 79.60 & \textcolor{red}{-0.74} $\vert$ 71.14 & \textcolor{red}{-5.61} $\vert$ 78.95 & \textcolor{red}{-0.04} $\vert$ 71.49 \\
        \hline
    \end{tabular}
    \caption{Channel selection using the l2-norm instead of distance between class prototypes.}
    \label{tab:l2}
\end{table}

The \textit{l2-norm} performs reasonably when combined with the three classifiers. Although the method enables ECP-Median-l2 with ROCKET to perform similar to the default version ECP-Median with euclidean distance, overall it is not better than the distance-based selection strategies. The reason for this could be that the \textit{l2-norm} of class-prototypes is highly susceptible to data noise for each channel, where a slight jitter could mislead results.
Hitherto, from Tables  \ref{tab:sotacs}, \ref{tab:madres} and \ref{tab:l2}, we conclude that distance-based selection strategies with the ECP-MAD as class prototype works best in comparison to other strategies.

\subsection{Sensitivity Analysis}
The above section shows that channel selection significantly reduces the computation time while maintaining accuracy. In this section, we take a more refined look into the data and analyse the performance of these techniques on the UEA datasets.

\subsubsection{Channel Utilisation for Different Class Prototypes}

\begin{figure}[!h]
    \centering
    \includegraphics[width=\textwidth]{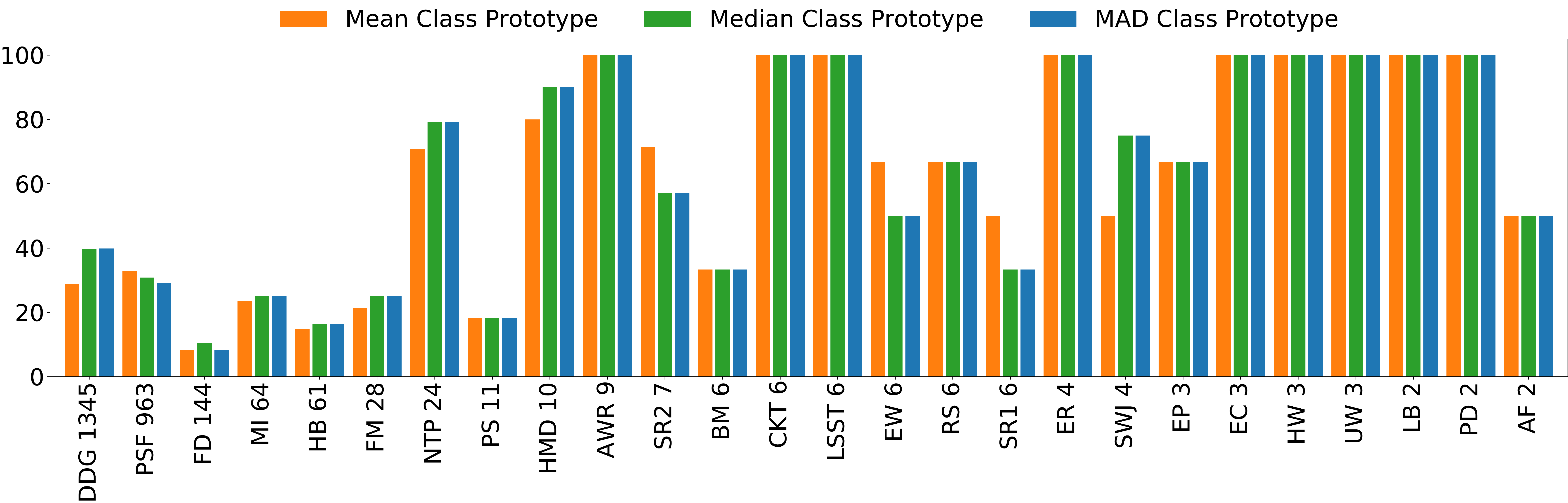}
    \caption{Channel utilisation for the three class prototypes with ECP selection.}
    \label{fig:channel_uti}
\end{figure}

Figure \ref{fig:channel_uti} shows the percentages of channels utilised by the three class prototypes. Interestingly, the higher the number of channels in the dataset, the fewer channels are helpful. Datasets like DuckDuckGeese has 1345 channels; however, only 538 channels are selected by ECP-MAD. As the number of channels decreases, more channels get utilised, which is logical as they tend to possess more information. 

\subsubsection{Runtime by Different Class Prototypes}
Table \ref{tab:timespace} presents time and memory required by different class prototypes along with various channel selection techniques. The time taken by MAD is highest among the three, as it iterates over data points for every class (as described in Section \ref{sec:meth}). The table also illustrates the memory required for different datasets. The proposed methods significantly reduce the memory required, as compared to the 1.2GB required to store the 26 UEA datasets.

\begin{table}[!h]
    \centering
    \begin{tabular}{cccc}
    \hline
    \multicolumn{1}{c}{Class Prototype$\rightarrow$}  &  \multicolumn{1}{c}{Mean} & \multicolumn{1}{c}{Median} & \multicolumn{1}{c}{MAD}\\
    \hline
      Channel Selection$\rightarrow$  & ECS $\vert$ ECP & ECS $\vert$ ECP & ECS $\vert$ ECP \\
    \hline
    Time (minutes) & 0.26 $\vert$ 0.27 & 0.31 $\vert$ 0.32 & 0.64 $\vert$ 0.67 \\
    Memory (GB) &  0.28 $\vert$ 0.42 & 0.27 $\vert$ 0.42  & 0.25 $\vert$ 0.41\\
    \hline
    \end{tabular}
    \caption{Total runtime of channel selection methods and the memory required to store the 26 UEA datasets after selection.}
    \label{tab:timespace}
\end{table}

\subsubsection{Analysis by Datasets}

As mentioned in Section \ref{subsec:data}, the UEA MTSC archive is a heterogeneous collection of problems. In this section, we discuss the performance of ECP on the benchmark datasets. 

\begin{figure}[H]
    \centering
    \includegraphics[width=\textwidth]{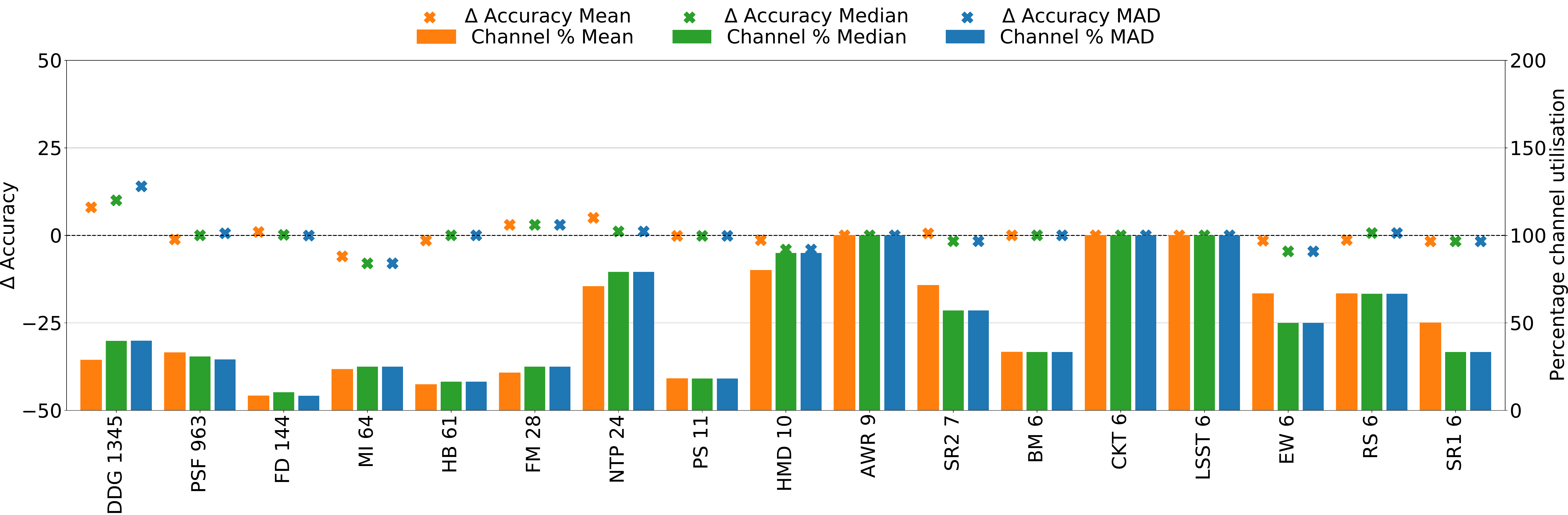}
    \caption{Rocket ECP accuracy and channel utilisation for three class-prototypes.}
    \label{fig:benchRocket}
\end{figure}

\begin{figure}[H]
    \centering
    \includegraphics[width=\textwidth]{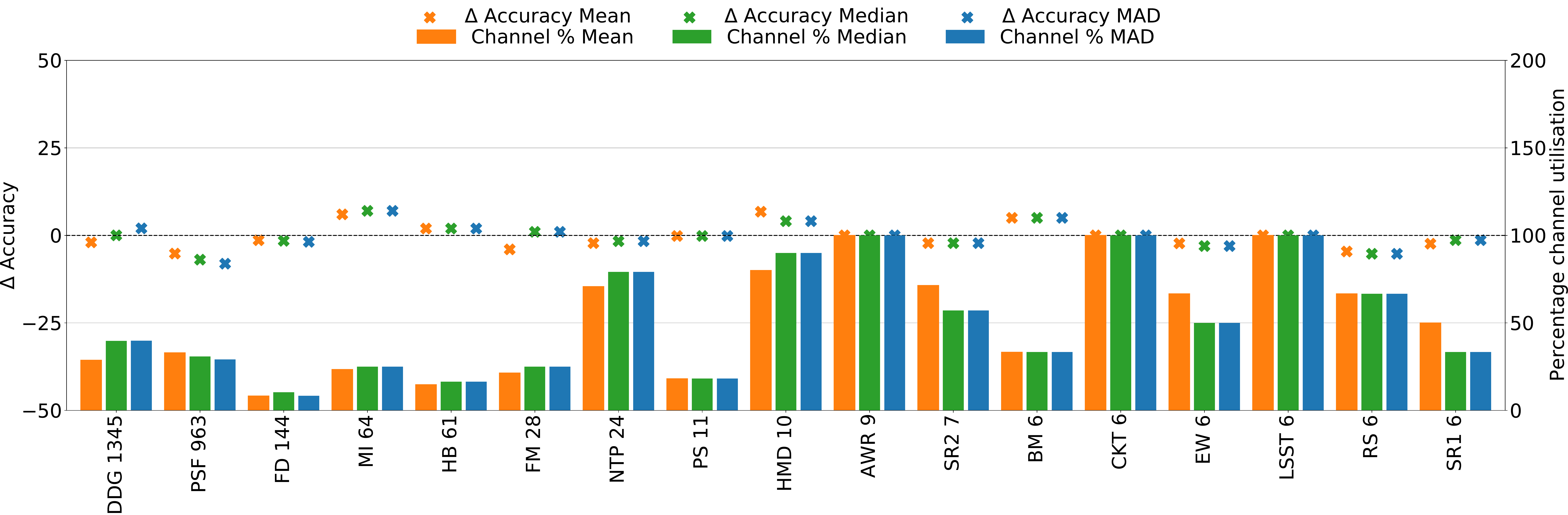}
    \caption{MrSEQL-SAX ECP accuracy and channel utilisation for three class-prototypes.}
    \label{fig:benchSEQL}
\end{figure}

\begin{figure}[H]
    \centering
    \includegraphics[width=\textwidth]{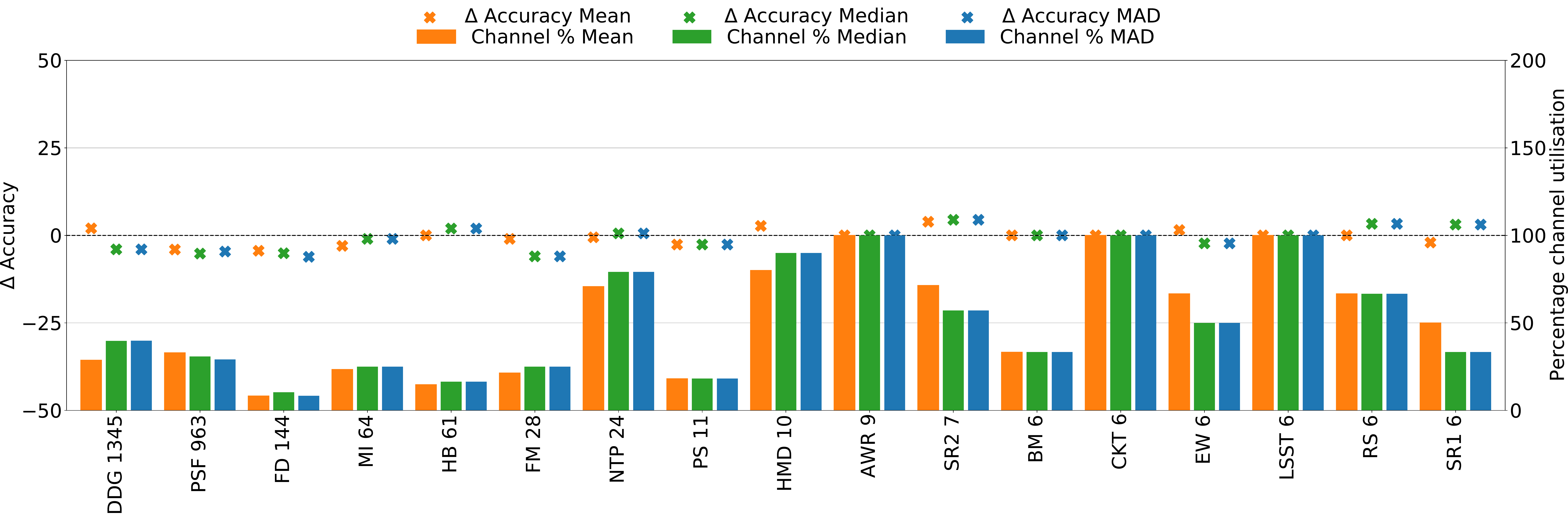}
    \caption{Weasel-Muse ECP accuracy and channel utilisation for three class-prototypes.}
    \label{fig:benchWM}
\end{figure}

Figure \ref{fig:benchRocket},\ref{fig:benchSEQL}  and \ref{fig:benchWM} illustrates the change in accuracy on different class prototypes versus the number of channels utilized by each dataset for ROCKET, MrSEQL and Weasel-Muse. 
We note that the accuracy is largely preserved, with a significant reduction in the number of channels, especially for datasets with more than 10 channels.
Although the set of channels recommended by the proposed techniques is identical for all classifiers, the performance of classifiers varies for the same set of datasets. The DuckDuckGeese (DDG) dataset gains significant accuracy for ROCKET and MrSEQL but loses some for Weasel-Muse. Similarly, for the MotorImagery( MI), MrSEQL-SAX performs better than the default, but ROCKET and Weasel-Muse are not as accurate. These results may arise due to the type of classifier, and the study of the classifier is outside the current study's scope. 

\subsubsection{Analysis by the Number of Channels}

The UEA datasets vary in the number of channels. It contains three datasets with more than 100 channels and six datasets with between 10 and 100 channels. Table \ref{tab:100} and Table \ref{tab:10} demonstrate the performance of channel selection for datasets in these groups. ROCKET outperforms the other two classifiers when the channels are greater than 100. It gains 4.48\% average accuracy on the three datasets at the cost of 13.8 (7.04\%) seconds more than the default setting. 

Although the channel selection for the other two classifiers does not help them as much as in the case of Rocket, it helps them reduce the time significantly. Moreover, the MrSEQL-SAX performs better than Rocket and Weasel Muse for datasets between 10 and 100 channels.

\begin{table}[!h]
    \centering
    \begin{tabular}{cccc}
    \hline
    \multicolumn{1}{c}{Class Prototype$\rightarrow$}  &  \multicolumn{1}{c}{Mean} & \multicolumn{1}{c}{Median} & \multicolumn{1}{c}{MAD}\\
    \hline
    Classifiers$\downarrow$ & $\Delta$Acc $\vert$ $\%$Time & $\Delta$Acc $\vert$ $\%$Time & $\Delta$Acc $\vert$ $\%$Time \\
    \hline
       Rocket  &  \textcolor{blue}{+2.60} $\vert$ 2.30 & \textcolor{blue}{+3.38} $\vert$ 1.13 & \textcolor{blue}{+4.84} $\vert$ -7.04\\
       WeaselMuse  & \textcolor{red}{-2.15} $\vert$ 76.52 & \textcolor{red}{-4.77} $\vert$ 74.33 & \textcolor{red}{-4.92} $\vert$ 79.06\\
       MrSEQL-SAX & \textcolor{red}{-2.86} $\vert$ 74.26 & \textcolor{red}{-2.83} $\vert$ 72.62 & \textcolor{red}{-2.65} $\vert$ 84.40\\
         \hline
    \end{tabular}
    \caption{ECP channel selection on datasets with channels $\ge$ 100}
    \label{tab:100}
\end{table}

\begin{table}[!h]
    \centering
    \begin{tabular}{cccc}
    \hline
    \multicolumn{1}{c}{Class Prototype$\rightarrow$}  &  \multicolumn{1}{c}{Mean} & \multicolumn{1}{c}{Median} & \multicolumn{1}{c}{MAD}\\
    \hline
    Classifiers$\downarrow$ & $\Delta$Acc $\vert$ $\%$Time & $\Delta$Acc $\vert$ $\%$Time & $\Delta$Acc $\vert$ $\%$Time \\
    \hline
       Rocket  & \textcolor{blue}{+0.08} $\vert$  20.96 & \textcolor{red}{-0.81} $\vert$ 20.27 & \textcolor{red}{-0.81} $\vert$ 20.47 \\
       WeaselMuse  & \textcolor{red}{-1.44} $\vert$ 76.32 & \textcolor{red}{-1.42} $\vert$  75.03& \textcolor{red}{-1.42} $\vert$ 84.40 \\
       MrSEQL-SAX & \textcolor{blue}{+0.31} $\vert$ 55.91 & \textcolor{blue}{+1.62} $\vert$ 55.74 & \textcolor{blue}{+1.62} $\vert$ 73.52\\
         \hline
    \end{tabular}
    \caption{ECP channel selection on datasets with channels $>$ 10 and $<$ 100}
    \label{tab:10}
\end{table}

\subsubsection{Analysis by Problem Domains}
The application of multivariate data varies across domains. In the UEA archive, the data comes from 5 different domains, namely, Audio Spectra Classification (ASC), EEG/MEG Classification (EEG), Human Activity Recognition (HAR), Motion Classification (MC), ECG Classification (ECG). 

Table \ref{tab:domain} illustrates the performance of the classifiers on the mentioned domains when using the ECP-MAD selection. Channel selection improves the performance of Rocket in ASR and MC domains, while in the case of EEG, MrSEQL performs slightly better with channel selection than its default setting. Channel selection also aids Weasel-Muse in the case of HAR and MC.   

\begin{table}[H]
    \centering
    \begin{tabular}{cccc}
    \hline
    Classifiers$\rightarrow$  & Rocket $\lvert$ MrSEQL $\lvert$ WM \\
    \hline
       ASR (3)   & \textcolor{blue}{+4.48} $\vert$ \textcolor{blue}{+1.26} $\vert$ \textcolor{red}{-1.55}\\
       ECG (2)  & \textcolor{red}{-4.00} $\vert$ \textcolor{red}{-6.67} $\vert$ \textcolor{red}{-6.66} \\
       EEG (6)   & \textcolor{red}{-0.28} $\vert$ \textcolor{blue}{+1.10} $\vert$ \textcolor{red}{-2.29}\\
    HAR (9) &   \textcolor{red}{-1.20} $\vert$ \textcolor{red}{-0.55} $\vert$ \textcolor{blue}{+0.18}\\
    MC (3) &  \textcolor{blue}{0.00}  $\vert$ \textcolor{red}{-0.24} $\vert$ \textcolor{blue}{+0.01} \\
    Other (3)  & \textcolor{blue}{+4.44} $\vert$ \textcolor{red}{-2.69} $\vert$ \textcolor{red}{-1.54}\\
         \hline
    \end{tabular}
    \caption{$\Delta$ Accuracy ECP-MAD channel selection for datasets from different domains.}
    \label{tab:domain}
\end{table}

\subsubsection{Analysis of Channel Selection on Minority Classes}
\label{sec:minclsuea}

Although most of the datasets in the UEA benchmark are balanced with regard to the number of samples in each class, we analyse the impact of channel selection on the imbalanced datasets in this section. The significantly imbalanced datasets  from the UEA benchmark are reported below with the F1 score (refer to Table \ref{tab:uea_cc} for class counts). Table \ref{tab:ueamin} compares ECP-MAD based channel selection (F1-CS) with the default classifier performance (F1-Default).  

\begin{table}[H]
    \centering
    \begin{tabular}{cccccc}
    \hline
    Dataset (\#Channels) &  \#Channels Selected & F1-Default & F1-CS & Best-Setting \\
    \hline
    EW (6)     &  4 & 83.36 & 86.58 & ECP+Mean\\
    HW (3)    &   3 & 55.72 & 55.72 & ECP+Med\\
    LSST (6)    & 6 & 43.66 & 43.66 & ECP+Med\\
    HB (61)  & 10 & 63.41 & 66.87 & ECS+MAD \\
    \hline
    \end{tabular}
    \caption{Analysis of channel-selection on imbalanced datasets.}
    \label{tab:ueamin}
\end{table}

The datasets EW, HW, LSST, and HB maintain or improve F1 scores with channel selection. For dataset HW and LSST, the channel selection method recommends using all channels; however, for EW and HB, it is recommended to use the fewer channels, especially HB sees a significant reduction in the number of channels to be used.
 Also, it should be noted that in most cases, ECP works well; however, the class prototypes are crucial for different types of channel selection (refer to Table \ref{tab:acc_aftercs}). 

%% file: 5-synth.tex
\section{Case Study 1: Synthetic Datasets}

The UEA benchmark suggests that our channel selection methods can reduce the computation time and memory while at least maintaining the same level of classification accuracy. However, other than using the resulting accuracy, we were unable to ascertain the effectiveness of the channel selection (i.e., did it select the right channels?) because we have no such ground truth. 

To address this problem, we make use of synthetic datasets from the work of \citep{NEURIPS2020_47a3893c}, where the important channels and timepoints are set during the data generation process.
We select the RareTime setting, where a small part of the time series is informative. The dataset type is generated using three signal types: Gaussian Process (GP), Pseudo Periodic (PP), and AutoRegressive (AR). The position of the signal can be stationary or moving (can change position within the channel). The length of time series is fixed to 100, and length of the important time series segment is 10. There are two classes.

\begin{figure}
\centering

\subfloat[]{\includegraphics[width=0.8\textwidth]{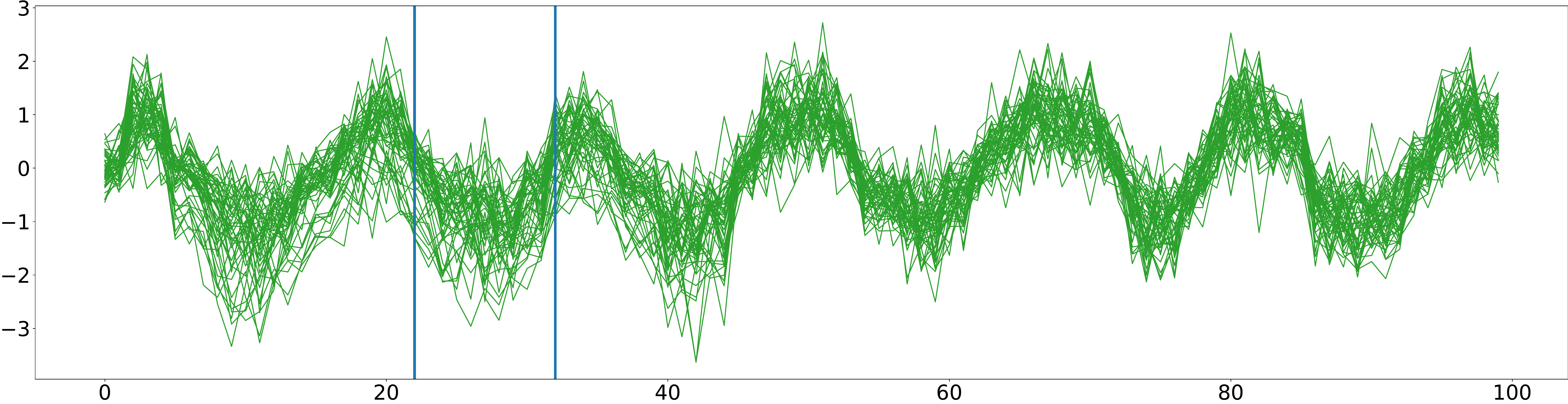}}

\subfloat[]{\includegraphics[width=0.8\textwidth]{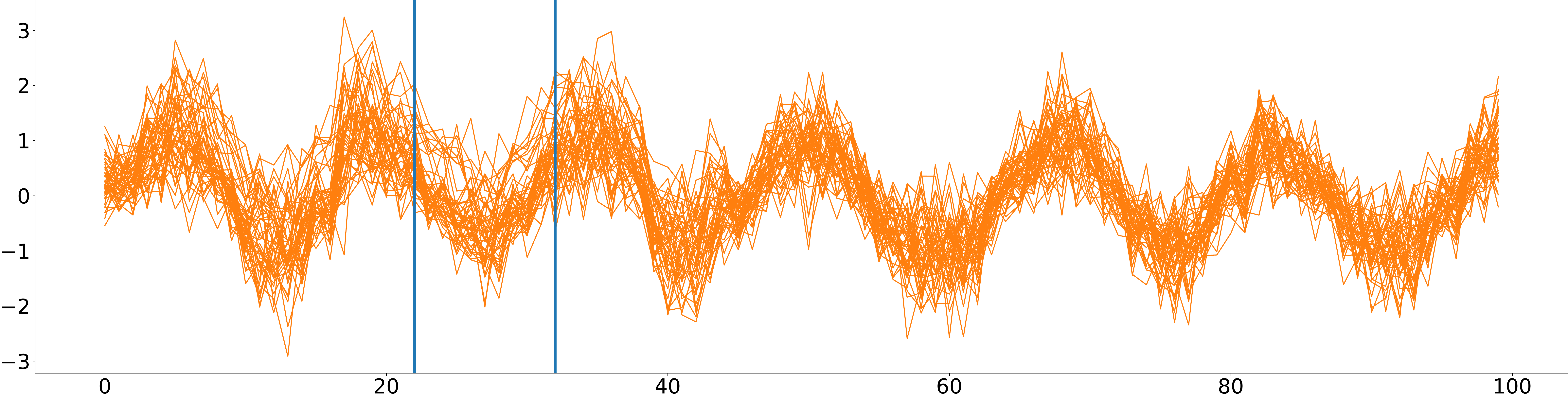}}
   
\caption[]{Two synthetic (multivariate) time series of different classes generated by the Pseudo Periodic process. The important time steps are 22-32 where the values in (a) are typically lower than (b).}
\label{fig:synthsample}
\end{figure}

\begin{figure}
    \centering
    \includegraphics[width=\textwidth]{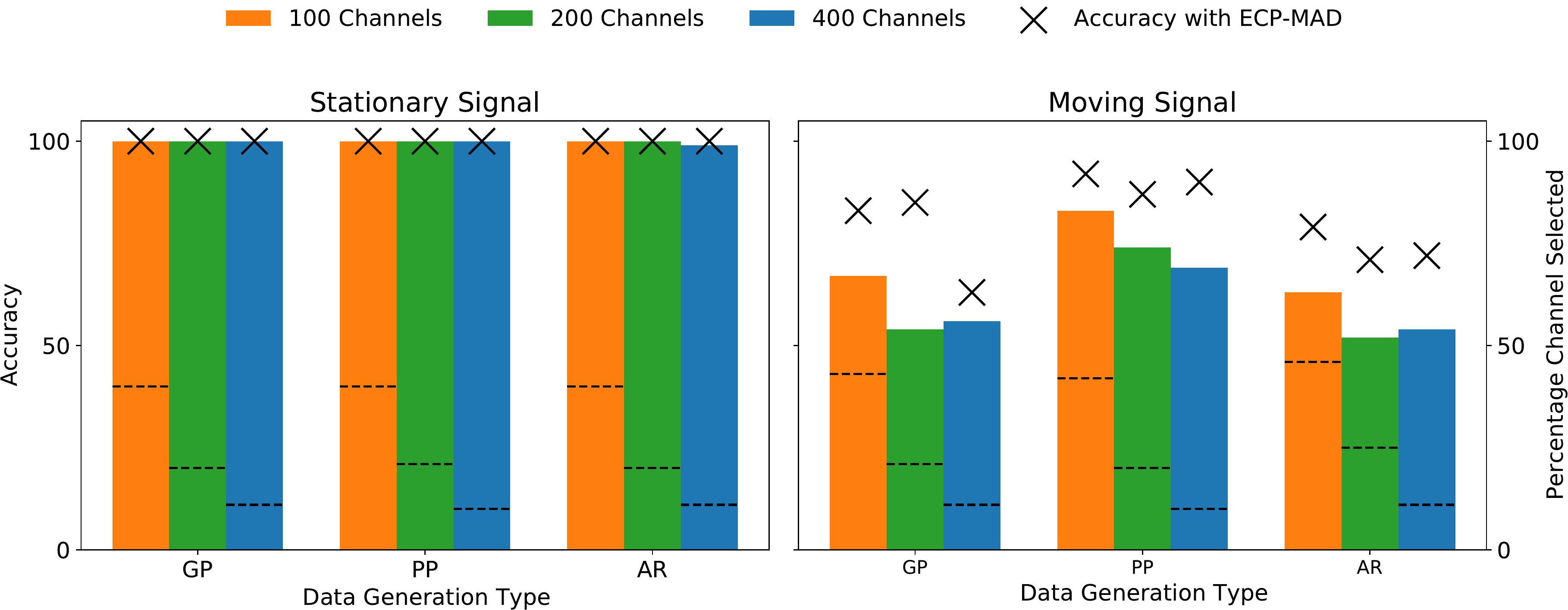}
    \caption{Synthetic datasets generated using the procedure in \citep{NEURIPS2020_47a3893c} using the RareTime setting. The dashed black line indicates the percentage of channels selected.}
    \label{fig:rare_time}
\end{figure}

The number of important channels is fixed to 40. Therefore, as the number of channels increases, the ratio of the important channel to all channels decreases. For example, from 40\%, it goes to 20\% and then 10\% for 100, 200 and 400 channels, so the problem becomes more difficult with an increasing number of channels.
Figure \ref{fig:rare_time} presents the performance of ROCKET on the synthetic data. ROCKET is the most accurate classifier among the three we studied; therefore, we use ROCKET to analyse the synthetic data. ROCKET achieves 100 percent accuracy for the stationary signal; however, it struggles to perform accurately when the signal is moving. When we use ROCKET with our channels selection method, ECP-MAD, then ROCKET achieves the same accuracy as without channel selection in the case of the stationary signal data; furthermore, in the case of moving signal, channel selection enables ROCKET to perform significantly better. 

The interesting point is that channel selection extracts approximately the same number of important channels mentioned during dataset generation. For 40\% of important channels in 100 channel stationary signal data, the channel selection recommends 40\% channels in static signal data for all three signals, and 43\%, 42\% and 46\% for moving signal for GP, PP, and AR signal data, respectively. Similarly, for 200 channel data, the important channels created are 40, the number of channels recommended is 20\%, 21\% and 20\% for stationary signal data and 21, 20 and 25 for moving signal data in GP, PP and AR, respectively. With the 400 channels, the number of important channels in the dataset reduces to 10\%, and it becomes harder for the classifier to find useful signal; however, the channel selection algorithm reduces the efforts required by the classifier. The method selects 11\%, 10\% and 11\% for stationary GP, PP and AR signal data and 14\%, 11\% and 13\% for moving GP, PP and AR signal data, respectively.

\subsection{Performance of Channel Selection on Minority Classes}

In continuation to Section \ref{sec:minclsuea}, we analyse the impact of channel selection on minority classes in synthetic data using ROCKET. Table \ref{tab:minority_syn} presents the F1 score for moving and stationary signals in GP, PP and AR datasets with 100 channels across the different scenarios of class imbalance. We vary the class ratio to evaluate the performance of channel selection. This enables us to study how the channel selection performs with different cardinalities. 

\begin{table}[!h]
    \centering
    \begin{tabular}{cccc}
    \hline
    Datasets & Class Ratio    &   \multicolumn{1}{c}{F1-Moving} & \multicolumn{1}{c}{F1-Stationary}\\
    \hline
     & \#0 : \#1 &  Default $\lvert$ CS &  Default  $\lvert$ CS \\
    \hline
    & 50:50 &  83.00 $\lvert$ 82.98 & 100 $\lvert$ 100\\
    GP & 90:10 &  34.21 $\lvert$ 34.21 & 81.00 $\lvert$ 97.99\\
    & 80:20 & 34.21 $\lvert$ 40.78 & 93.93 $\lvert$ 100\\
    \hline
    & 50:50 &  92.00 $\lvert$  92.00 & 100 $\lvert$ 100\\
    PP & 90:10 &  33.33 $\lvert$ 35.51 & 100 $\lvert$ 100 \\
    & 80:20 & 35.52 $\lvert$ 61.55 & 100 $\lvert$ 100\\
    \hline
    & 50:50 &   79.00 $\lvert$ 78.64 & 100 $\lvert$ 100 \\
    AR & 90:10 &  29.07 $\lvert$ 29.07 & 77.52 $\lvert$ 94.97\\
    & 80:20 & 29.07 $\lvert$ 34.54 & 97.99 $\lvert$ 98.99\\
    \hline
    \end{tabular}
    \caption{Analysis of ECP-MAD channel selection in minority class on synthetic data.}
    \label{tab:minority_syn}
\end{table}

Similar to the analysis above, we find that the cardinality of a class does not impact the channel selection. However, with various levels of imbalance, even a good classifier like ROCKET struggles to perform better. The channel selection, on some occasions, improves accuracy; however, not as much as in the case of balanced data.

%% file: 6-usecase.tex
\section{Case Study 2: Military Press Dataset}
\label{sec:case_study}

The experiments in the above sections confirm that our channel selection methods enable the state-of-the-art classifiers to achieve better accuracy in significantly  smaller amounts of time and memory and, in the worst case, saves time and memory with minimal loss of accuracy. However, to perform a sanity check on real-world data, we provide an additional study on Military Press data, collected and verified at University College Dublin with the help of domain experts.

\subsection{Dataset}

A total of 56 healthy volunteers (34 males and 22 females; age: 26 $\pm$ 5 years, height: 1.73 $\pm$ 0.09 m and body mass: 72 $\pm$ 15 kg) participated in a study aimed at analysing the execution of the Military Press strength and conditioning exercise. 
The participants completed ten repetitions of the normal form and ten repetitions of induced forms. The NSCA guidelines were applied under the guidance of sports physiotherapists and conditioning coaches to ensure standardisation. The dataset was extracted from the video of individuals performing the exercise with the help of the human body pose estimation library  OpenPose\footnote{\url{https://github.com/CMU-Perceptual-Computing-Lab/openpose}}. There are four classes in the dataset, namely: Normal (N), Asymmetrical (A), Reduced Range (R) and Arch (arch). The N refers to the correct execution of the exercise; A refers to when the barbell is lopsided and asymmetrical, R refers to the form where the bar is not brought down completely to the shoulder level and Arch refers to when participants arch their back. A total of 25 body parts were tracked along X and Y axis, as seen in Figure \ref{fig:elbowcut}. These 50 body trackers (25$\times$2) act as channels for the MTSC task. The train and test size for this dataset is 1452 and 601 respectively and the length of time-series is 160. 

\subsection{Evaluation}

When used with all 50 channels, ROCKET takes 1.45 minutes and achieves an accuracy of 77.53\%. In this setting, the normalisation hyperparameter of ROCKET is switched off; this is because the signal magnitude contains crucial information. 
Table \ref{tab:mtsc_cs_mp} shows accuracy and computation results before and after channel selection. 
\begin{table}[!h]
    \centering
    \begin{tabular}{cccc}
    \hline
              & \multicolumn{1}{c}{Before Selection} &  \multicolumn{1}{c}{After Selection} &\\
              \hline
    Classifiers $\downarrow$      &  Acc $\vert$ Time $\vert$ Mem (MB) & Acc $\vert$ Time $\vert$ Mem (MB) & Best Setting \\
    \hline
    ROCKET & 77.53 $\vert$ 1.45 $\vert$  98.5 & 84.02 $\vert$ 1.52 $\vert$ 13.80 & ECS-Median \\
    Weasel Muse & 57.57 $\vert$ 71.55 $\vert$ 98.5  & 59.90 $\vert$ 38.87 $\vert$  43.38 & ECP-Median\\
    MrSEAL-SAX  & 61.56 $\vert$ 633.56 $\vert$ 98.5   & 59.07 $\vert$ 360.80 $\vert$  43.38 & ECP-Median  \\
    \hline
    \end{tabular}
    \caption{The SOTA classifiers on 50 Channel Military Press dataset.}
    \label{tab:mtsc_cs_mp}
\end{table}
\subsection{Performance of SOTA Classifiers on MP Dataset}

Table \ref{tab:mpcs} demonstrates the performance of channel selection on the Military Press dataset. The ECS with Median as class prototype outperforms any other setting; the gain in accuracy is about 7\%. All the other channel selection methods perform better than ROCKET in its default setting.

\begin{table}[!h]
    \centering
    \begin{tabular}{c|c|c|c|c}
    \hline
    \multicolumn{1}{c|}{Center$\rightarrow$}  & \multicolumn{2}{c}{Median} &  \multicolumn{2}{c}{MAD}\\
    \hline
         Channel Selection$\rightarrow$ &  ECS & ECP & ECS & ECP \\
         \hline
        Classifier$\downarrow$            & Acc $\vert$ Time & Acc $\vert$ Time & Acc $\vert$ Time & Acc $\vert$ Time \\
        \hline
        ROCKET &  84.02 $\vert$ 1.52  &   81.03 $\vert$ 1.78  &  82.59 $\vert$ 1.59   &  80.20 $\vert$ 1.75  \\
        Weasel Muse &   54.57 $\vert$ 10.28 &  59.90 $\vert$ 38.87 & 54.57 $\vert$ 10.23 & 58.57 $\vert$ 48.35 \\
        MrSEQL-SAX &  58.90 $\vert$ 58.63 & 59.07 $\vert$ 360.80 & 58.90 $\vert$ 58.63 & 51.32 $\vert$ 382.08   \\
        \hline
    \end{tabular}
    \caption{Performance of SOTA classifiers on MP dataset.}
    \label{tab:mpcs}
\end{table}

\textbf{Ranking:} The channel ranking using the best method here, ECS with median prototype, is as follows: 
'RWrist\_Y', 'LWrist\_Y', 'RElbow\_Y', 'LElbow\_Y', 'RElbow\_X', 'LElbow\_X', 'RHeel\_Y'. We note that out of 50 channels, this method selects only a subset of 7 channels, which not only reduces the data drastically, but also results in significantly increased accuracy. We consider this a very good result, also considering that it makes it easier to provide feedback to the participant executing the exercise, by pointing to the most important body parts for the execution.

\subsection{Performance of Channel Selection with Two Classes}

With the help of domain experts, we know that the Military Press exercises with classes Asymmetric, Reduced, and Arch, are exercises done incorrectly. All three classes have different signal magnitudes. To study the impact of channel selection when one of the classes has a group of different behaviours, we group the classes mentioned above into Abnormal and train a binary classifier, Abnormal, against the Normal class in the dataset. Since there are only two classes, ECS and ECP strategies are identical. 

\begin{table}[!h]
    \centering
    \begin{tabular}{cccc}
    \hline
    All & Mean & Median & MAD \\
    \hline
    72.05 & 73.71 & 77.05 & 77.50 \\
    \hline
    \end{tabular}
    \caption{F1 score for ROCKET when using two classes}
    \label{tab:mp_normvsab}
\end{table}

Table \ref{tab:mp_normvsab} illustrates the performance of ROCKET on the three prototypes used. We see that channel selection helps ROCKET to distinguish between the two classes. Furthermore, MAD is the most effective strategy. Since this is an imbalanced dataset, we also compared precision and recall in Table \ref{tab:mpcr} and found that, with channel selection, there is an improvement in both precision and recall.

\begin{table}[!htb]

    \begin{minipage}{.5\linewidth}
      \centering
        \begin{tabular}{l|lll}
        \hline
             & Precision & Recall & F1\\
        \hline
             Abnormal &  0.86 & 0.86 & 0.86 \\
             Normal & 0.59 & 0.57 & 0.58 \\
        \hline
        
        \end{tabular}
        \caption*{F1 results with all channels.}
    \end{minipage}%
    \begin{minipage}{.5\linewidth}
      \centering
        
        \begin{tabular}{lll|l}
        \hline
            Precision & Recall & F1 & Support \\
        \hline    
            0.90 & 0.87 & 0.88 & 451\\
            0.64 & 0.69 & 0.67 & 150 \\
        \hline
        \end{tabular}
        \caption*{F1 results for the MAD prototype.}
        
    \end{minipage} 
  
    \caption{Comparison of performance of ROCKET with and without channel selection.}
    \label{tab:mpcr}
\end{table}

%% file: 8-limit.tex
\section{Limitations \& Future Work}
\label{sec:limit}

While ECP performs better than ECS in most datasets, it may collect some noisy channels when iterating through the different class pairs in the distance matrix. These noisy channels, along with useful channels, are then used for classification and this could negatively impact the classifier accuracy.
The proposed methods rely on the euclidean distance between two class prototypes; however, if a class prototype is too noisy, it could amplify the euclidean distance. With the help of the signal's magnitude, the euclidean distance could be regularised. In our future work we want to further analyse the impact of combining information from the prototype magnitude and the distance-based computation to address this issue.

We also plan to investigate whether combining multiple class prototypes can further improve results. At the moment we only use one class prototype to represent each class and for classes with subgroups of behaviours this could be an issue. We can consider first clustering the behaviours in each class and representing each subgroup with a prototype. In that case, we also need to investigate whether all class prototypes are equally important, or if some are more important than others.

The current study focuses on channel selection based on euclidean distance and does not use time-series attributes or statistical properties of each class to adjust the selection strategy. Similar to channel selection, we could also explore the time series segments that contribute most to the classification, providing another avenue to make existing MTSC algorithms more scalable.

%% file: 9-conclusion.tex
\section{Conclusion}

This study shows that for many MTSC datasets, state-of-the-art classifiers do not require all the channels to achieve the best accuracy. Some channels are very noisy for the classification task and act as a bottleneck for the classifier to achieve its maximum potential. We observed that our channel selection methods remove data noise and drastically reduce the required computation time for existing MTSC methods. The current study showed that the distance between the class prototypes of various channels plays a crucial role in identifying the noisy channels. Our channel selection strategies, ECP and ECS, can select the useful channels based on euclidean distance between class prototypes. Our techniques significantly reduced the runtime and memory required to run MTSC classifiers. There is some accuracy loss for some datasets, however, this is minimal, and the benefit of data reduction and computation savings outweights the small accuracy loss.
Channel selection saves approximately 75\% time and around 60\% memory. The proposed channel selection techniques have more impact on the datasets wiht a higher number of channels. We verified the performance of our channel selection methods on various types of signals with varying levels of difficulty, ranging from the popular UEA MTSC benchmark, to synthetic data and a real-world case study. The channels selected performed very well on those datasets, enhancing the performance of the most accurate classifier, ROCKET. On our case study with a real-world dataset, Military Press, our channel selection methods combined with ROCKET improved the accuracy by more than 5\%. The proposed channel selection techniques are classifier-agnostic, they can be plugged into any classification pipeline, before training an MTSC classifier. They are also  simple, fast and at the same time robust to different types of noise, which leads to preserved classifier accuracy as shown in our extensive experiments.

%% file: appendix.tex
\section*{Appendix}

\subsection*{Evaluation metrics}

\[
\text{Accuracy}  = \frac{\text{Number of correct predictions}}{\text{Total number of predictions}}
\]

\[
\text{F1 score} = 2 * \frac{\text{Precision * Recall}}{\text{Precision + Recall}}
\]

\[
\Delta \text{Acc} = \text{Accuracy after channel selection} - \text{Accuracy with default classifier} 
\]

\[
\% \text{Time} = \frac{\text{Time taken with Channel Selection}- \text{Time taken without Channel Selection}}{\text{Time taken without Channel Selection}} 
\]

\begin{table}[!h]
    \centering
\begin{tabular}{cccccccc}
\rot{Domain} & \rot{Problem} & \rot{Acronym} & \rot{Train Size} & \rot{Test Size} & \rot{Num Channels} & \rot{Series Length} & \rot{Num Classes} \\
\hline
ASC  & DuckDuckGeese & DDG & 50 & 50 & 1345 & 270 & 5 \\
Other & PEMS-SF & PSF & 267 & 173 & 963 & 144 & 7 \\
EEG  & FaceDetection & FD & 5890 & 3524 & 144 & 62 & 2 \\
EEG  & MotorImagery & MI & 278 & 100 & 64 & 3000 & 2 \\
ASC  & Heartbeat & HB & 204 & 205 & 61 & 405 & 2 \\
EEG  & FingerMovements & FM & 316 & 100 & 28 & 50 & 2 \\
HAR & NATOPS & NTP & 180 & 180 & 24 & 51 & 6 \\
ASC  & PhonemeSpectra & PS & 3315 & 3353 & 11 & 217 & 39 \\
EEG  & HandMovementDirection & HMD & 160 & 74 & 10 & 400 & 4 \\
Mot  & ArticularyWordRecognition & AWR & 275 & 300 & 9 & 144 & 25 \\
EEG  & SelfRegulationSCP2 & SR2 & 200 & 180 & 7 & 1152 & 2 \\
HAR & BasicMots & BM & 40 & 40 & 6 & 100 & 4 \\
HAR & Cricket & CKT & 108 & 72 & 6 & 1197 & 12 \\
HAR & EigenWorms & EW & 128 & 131 & 6 & 17984 & 5 \\
HAR & LSST & LSST & 2459 & 2466 & 6 & 36 & 14 \\
HAR & RacketSports & RS & 151 & 152 & 6 & 30 & 4 \\
EEG  & SelfRegulationSCP1 & SR1 & 268 & 293 & 6 & 896 & 2 \\
HAR & ERing & ER & 30 & 270 & 4 & 65 & 6 \\
ECG  & StandWalkJump & SWJ & 12 & 15 & 4 & 2500 & 3 \\
Mot  & Epilepsy & EP & 137 & 138 & 3 & 206 & 4 \\
Other & EthanolConcentration & EC & 261 & 263 & 3 & 1751 & 4 \\
HAR & Handwriting & HW & 150 & 850 & 3 & 152 & 26 \\
HAR & UWaveGestureLibrary & UW & 120 & 320 & 3 & 315 & 8 \\
ECG  & AtrialFibrillation & AF & 15 & 15 & 2 & 640 & 3 \\
Other & Libras & LB & 180 & 180 & 2 & 45 & 15 \\
Mot  & PenDigits & PD & 7494 & 3498 & 2 & 8 & 10 \\
\hline

\end{tabular}
    \caption{Data Dictionary for 26 equal length UCR datasets.}
    \label{tab:data_dic}
\end{table}

\begin{table}[]
    \centering
\begin{tabular}{c|cccccc}
\hline
Channels $\downarrow$ & \multicolumn{4}{c}{$\leftarrow$ Classes $\rightarrow$}\\  
 & CP\_a\_arch & CP\_a\_n & CP\_a\_r & CP\_arch\_n & CP\_arch\_r & CP\_n\_r \\
\hline
RWrist\_Y & 51.00 & 56.27 & 103.15 & 48.93 & 123.44 & 82.84 \\
RElbow\_Y & 50.66 & 51.67 & 98.55 & 35.13 & 87.03 & 75.97 \\
LWrist\_Y & 40.12 & 45.61 & 99.03 & 45.41 & 123.17 & 90.40 \\
REar\_Y & 39.44 & 5.29 & 11.39 & 36.79 & 31.18 & 9.85 \\
LEar\_Y & 38.68 & 3.26 & 4.04 & 39.84 & 37.40 & 5.99 \\
RShoulder\_Y & 35.43 & 11.63 & 23.36 & 34.42 & 18.33 & 23.77 \\
RElbow\_X & 30.66 & 43.73 & 76.51 & 30.60 & 70.06 & 68.05 \\
RHip\_Y & 28.88 & 6.41 & 11.38 & 27.95 & 19.21 & 10.52 \\
MidHip\_Y & 28.23 & 8.10 & 9.16 & 25.60 & 21.45 & 7.49 \\
LElbow\_Y & 27.72 & 40.91 & 95.98 & 33.26 & 92.82 & 83.06 \\
LElbow\_X & 27.51 & 40.61 & 55.97 & 28.79 & 68.01 & 68.89 \\
LShoulder\_Y & 27.22 & 16.11 & 10.78 & 37.82 & 25.27 & 20.57 \\
Neck\_Y & 25.91 & 14.47 & 15.41 & 36.49 & 18.38 & 25.95 \\
RSmallToe\_X & 25.09 & 24.26 & 29.78 & 32.07 & 33.48 & 36.95 \\
RWrist\_X & 24.63 & 19.88 & 17.88 & 14.93 & 29.84 & 28.10 \\
LHip\_Y & 23.97 & 9.02 & 7.78 & 23.49 & 19.12 & 8.11 \\
RBigToe\_Y & 20.66 & 20.10 & 16.96 & 29.57 & 25.34 & 30.02 \\
RSmallToe\_Y & 20.05 & 14.86 & 12.26 & 26.29 & 19.48 & 20.93 \\
Neck\_X & 17.27 & 6.37 & 20.87 & 15.78 & 26.78 & 22.75 \\
MidHip\_X & 16.87 & 10.10 & 11.46 & 13.77 & 15.98 & 9.30 \\
LBigToe\_X & 13.31 & 6.06 & 5.71 & 13.76 & 11.98 & 6.10 \\
LWrist\_X & 13.27 & 10.90 & 14.93 & 9.98 & 15.47 & 10.37 \\
LKnee\_X & 11.57 & 7.31 & 9.04 & 11.69 & 14.93 & 8.73 \\
LAnkle\_X & 11.02 & 6.89 & 10.52 & 10.02 & 15.16 & 12.29 \\
RBigToe\_X & 10.45 & 6.22 & 5.57 & 11.79 & 11.26 & 8.36 \\
RHeel\_Y & 10.40 & 1.90 & 4.40 & 10.55 & 12.48 & 4.58 \\
LSmallToe\_Y & 10.10 & 7.58 & 11.15 & 9.87 & 10.42 & 10.03 \\
LHip\_X & 9.59 & 7.26 & 6.46 & 10.97 & 13.94 & 9.65 \\
RKnee\_X & 9.54 & 8.88 & 6.95 & 12.40 & 11.68 & 7.27 \\
REar\_X & 8.92 & 16.58 & 7.71 & 16.50 & 7.16 & 14.55 \\
LHeel\_Y & 8.81 & 5.90 & 6.90 & 6.53 & 10.84 & 6.74 \\
LKnee\_Y & 7.69 & 6.18 & 6.07 & 6.77 & 7.70 & 4.85 \\
RAnkle\_Y & 7.60 & 1.19 & 6.49 & 7.36 & 9.31 & 6.41 \\
LEar\_X & 7.09 & 2.14 & 3.22 & 6.27 & 6.74 & 2.59 \\
RShoulder\_X & 6.87 & 10.69 & 6.45 & 10.55 & 7.39 & 12.42 \\
LBigToe\_Y & 6.85 & 6.74 & 6.54 & 5.68 & 5.50 & 3.37 \\
RHip\_X & 6.43 & 7.97 & 6.47 & 3.46 & 1.20 & 3.89 \\
LShoulder\_X & 6.38 & 20.07 & 11.29 & 20.81 & 13.67 & 20.76 \\
RKnee\_Y & 6.20 & 5.50 & 6.16 & 4.06 & 5.35 & 4.75 \\
LSmallToe\_X & 5.89 & 27.39 & 7.29 & 29.62 & 7.87 & 28.84 \\
RAnkle\_X & 4.28 & 2.47 & 0.15 & 5.06 & 4.28 & 2.47 \\
RHeel\_X & 3.99 & 2.20 & 0.12 & 4.90 & 4.02 & 2.19 \\
LAnkle\_Y & 0.44 & 1.78 & 3.62 & 1.75 & 3.66 & 4.03 \\
LHeel\_X & 0.16 & 16.05 & 13.26 & 16.08 & 13.29 & 17.56 \\
Nose\_X & 0.00 & 0.00 & 0.00 & 0.00 & 0.00 & 0.00 \\
REye\_X & 0.00 & 0.00 & 0.00 & 0.00 & 0.00 & 0.00 \\
LEye\_X & 0.00 & 0.00 & 0.00 & 0.00 & 0.00 & 0.00 \\
Nose\_Y & 0.00 & 0.00 & 0.00 & 0.00 & 0.00 & 0.00 \\
REye\_Y & 0.00 & 0.00 & 0.00 & 0.00 & 0.00 & 0.00 \\
LEye\_Y & 0.00 & 0.00 & 0.00 & 0.00 & 0.00 & 0.00 \\

\hline
\end{tabular}
    \caption{Distance matrix for the Median class-prototype on the MP dataset with 50 channels.}
    \label{tab:dm_mad}
\end{table}

\begin{landscape}

\begin{table}[]
    \centering
\begin{tabular}{c|ccc|ccc|cc|ccc}
\hline
& \multicolumn{3}{c|}{Before Channel Selection}& \multicolumn{3}{c|}{After Channel Selection} & \multicolumn{2}{c|}{Best classifier \& Accuracy}\\
\hline
\rot{Dataset} & \rot{Rocket} & \rot{WM} & \rot{SEQL} & \rot{CS Rocket} & \rot{CS WM} & \rot{CS SEQL} & \rot{Best Classifier} & \rot{Best Accuracy} & \rot{$\Delta$ Acc} & \rot{Best CS} & \rot{Best CP}\\
\hline
DDG 1345 & 46.00 & 44.00 & 34.00 & 60.00 & 50.00 & 42.00 & Rocket & 60.00 & 14.00 & ECP & MAD \\
PSF 963 & 84.39 & 98.27 & 95.95 & 84.97 & 94.22 & 90.75 & WM & 94.22 & -4.05 & ECP & MEAN \\
FD 144 & 62.74 & 65.21 & 55.90 & 63.71 & 60.81 & 54.51 & Rocket & 63.71 & 0.97 & ECP & MED \\
MI 64 & 57.00 & 55.00 & 51.00 & 51.00 & 63.00 & 58.00 & WM & 63.00 & 8.00 & ECS & MEAN \\
HB 61 & 74.15 & 73.17 & 73.17 & 74.15 & 75.12 & 75.12 & WM & 75.12 & 1.95 & ECP & MED \\
FM 28 & 53.00 & 54.00 & 56.00 & 56.00 & 60.00 & 57.00 & WM & 60.00 & 6.00 & ECS & MEAN \\
NTP 24 & 87.22 & 91.67 & 87.22 & 92.22 & 92.22 & 85.56 & Rocket & 92.22 & 5.00 & ECP & MED \\
PS 11 & 27.62 & 31.46 & 26.27 & 27.47 & 29.29 & 26.10 & WM & 29.29 & -2.17 & ECS & MED \\
HMD 10 & 54.05 & 28.38 & 14.86 & 54.05 & 31.08 & 22.97 & Rocket & 54.05 & 0.00 & ECP & MED \\
AWR 9 & 99.33 & 99.33 & 99.33 & 99.33 & 99.33 & 99.33 & Rocket & 99.33 & 0.00 & ECP & MED \\
SR2 7 & 53.89 & 47.78 & 50.56 & 55.00 & 56.11 & 48.33 & WM & 56.11 & 8.33 & ECS & MEAN \\
BM 6 & 100.00 & 100.00 & 95.00 & 100.00 & 100.00 & 100.00 & Rocket & 100.00 & 0.00 & ECP & MED \\
CKT 6 & 100.00 & 100.00 & 98.61 & 100.00 & 100.00 & 98.61 & Rocket & 100.00 & 0.00 & ECP & MED \\
EW 6 & 54.62 & 89.31 & 73.28 & 54.62 & 90.84 & 70.99 & WM & 90.84 & 1.53 & ECP & MEAN \\
LSST 6 & 89.31 & 61.27 & 58.80 & 87.79 & 61.27 & 58.80 & Rocket & 87.79 & -1.52 & ECP & MED \\
RS 6 & 90.79 & 86.18 & 86.84 & 91.45 & 89.47 & 82.24 & Rocket & 91.45 & 0.66 & ECP & MED \\
SR1 6 & 84.64 & 78.50 & 68.26 & 82.94 & 81.57 & 66.89 & Rocket & 82.94 & -1.70 & ECP & MED \\
ER 4 & 98.15 & 96.67 & 87.78 & 98.15 & 96.67 & 87.78 & Rocket & 98.15 & 0.00 & ECP & MED \\
SWJ 4 & 53.33 & 46.67 & 33.33 & 53.33 & 60.00 & 40.00 & WM & 60.00 & 13.33 & ECP & MED \\
EP 3 & 98.55 & 100.00 & 99.28 & 100.00 & 100.00 & 98.55 & Rocket & 100.00 & 1.45 & ECP & MED \\
EC 3 & 43.35 & 36.50 & 55.51 & 49.81 & 44.49 & 58.17 & SEQL & 58.17 & 2.66 & ECP & MAD \\
HW 3 & 59.53 & 26.12 & 47.41 & 59.53 & 26.12 & 47.41 & Rocket & 59.53 & 0.00 & ECP & MED \\
UW 3 & 94.06 & 90.63 & 87.19 & 94.06 & 90.63 & 87.19 & Rocket & 94.06 & 0.00 & ECP & MED \\
AF 2 & 6.67 & 40.00 & 26.67 & 20.00 & 13.33 & 40.00 & SEQL & 40.00 & 13.33 & ECS & MED \\
LB 2 & 90.56 & 93.33 & 87.22 & 90.56 & 93.33 & 87.22 & WM & 93.33 & 0.00 & ECP & MED \\
PD 2 & 98.20 & 93.85 & 92.28 & 98.20 & 93.85 & 92.28 & Rocket & 98.20 & 0.00 & ECP & MED \\
\hline
\end{tabular}

    \caption{Accuracy performance of channel selection on UEA MTSC datasets with different channel selection and channel prototypes.}
    \label{tab:acc_aftercs}
\end{table}
\end{landscape}

\newpage

\begin{landscape}
\begin{table}[]
    \centering
    \begin{tabular}{c|ccc|ccc|cc}
    \hline
    & \multicolumn{3}{c|}{Time before channel selection}  & \multicolumn{3}{c|}{Time after channel selection} & Storage\\
\hline
\rot{Dataset} & \rot{Time Rocket} & \rot{Time WM} & \rot{Time SEQL} & \rot{Best Time Rocket} & \rot{Best Time WM} & \rot{Best Time SEQL} & \rot{Full Storage} & \rot{Best storage} \\
\hline
DDG 1345 & 0.13 & 302.52 & 249.73 & 0.20 & 40.58 & 106.66 & 147.25 & 58.68 \\
PSF 963 & 0.33 & 621.38 & 688.13 & 0.43 & 229.96 & 491.53 & 315.83 & 104.29 \\
FD 144 & 2.88 & 987.31 & 3979.68 & 2.81 & 102.13 & 650.56 & 511.20 & 53.25 \\
MI 64 & 5.90 & 928.27 & 730.60 & 5.95 & 212.61 & 317.84 & 409.53 & 95.98 \\
HB 61 & 0.57 & 95.55 & 37.13 & 0.55 & 30.26 & 22.32 & 40.06 & 6.57 \\
FM 28 & 0.11 & 10.38 & 4.00 & 0.10 & 5.12 & 6.44 & 4.52 & 0.97 \\
NTP 24 & 0.06 & 6.07 & 2.58 & 0.07 & 10.60 & 9.56 & 2.24 & 1.77 \\
PS 11 & 4.94 & 309.61 & 990.72 & 2.56 & 58.13 & 448.86 & 65.10 & 11.84 \\
HMD 10 & 0.43 & 11.40 & 13.18 & 0.42 & 17.78 & 32.61 & 5.09 & 4.58 \\
AWR 9 & 0.26 & 8.68 & 9.30 & 0.26 & 17.47 & 35.42 & 3.04 & 3.04 \\
SR2 7 & 1.41 & 37.63 & 16.78 & 1.12 & 41.00 & 33.81 & 12.49 & 8.92 \\
BM 6 & 0.02 & 0.54 & 0.18 & 0.02 & 0.44 & 0.15 & 0.21 & 0.07 \\
CKT 6 & 0.74 & 16.15 & 16.61 & 0.75 & 28.20 & 53.75 & 6.00 & 6.00 \\
EW 6 & 0.54 & 344.38 & 511.94 & 0.54 & 207.33 & 476.45 & 105.47 & 6.00 \\
LSST 6 & 13.73 & 11.60 & 15.39 & 10.25 & 22.61 & 49.92 & 5.97 & 5.97 \\
RS 6 & 0.03 & 0.54 & 0.11 & 0.03 & 1.02 & 0.44 & 0.32 & 0.22 \\
SR1 6 & 1.37 & 36.05 & 14.47 & 0.84 & 25.17 & 19.56 & 11.20 & 3.73 \\
ER 4 & 0.01 & 0.90 & 0.09 & 0.01 & 2.46 & 0.50 & 0.08 & 0.08 \\
SWJ 4 & 0.15 & 2.88 & 1.54 & 0.07 & 4.89 & 3.39 & 0.92 & 0.69 \\
EP 3 & 0.12 & 1.94 & 0.72 & 0.10 & 3.13 & 2.16 & 0.70 & 0.47 \\
EC 3 & 1.96 & 31.19 & 25.89 & 0.97 & 21.66 & 29.92 & 10.56 & 10.56 \\
HW 3 & 0.10 & 4.42 & 1.85 & 0.11 & 10.00 & 9.35 & 0.58 & 0.58 \\
UW 3 & 0.16 & 3.57 & 2.16 & 0.16 & 7.98 & 10.05 & 0.91 & 0.91 \\
AF 2 & 0.03 & 0.44 & 0.24 & 0.02 & 0.47 & 0.60 & 0.15 & 0.08 \\
LB 2 & 0.03 & 0.37 & 0.20 & 0.04 & 1.07 & 1.22 & 0.17 & 0.17 \\
PD 2 & 0.91 & 0.71 & 0.82 & 0.90 & 2.05 & 5.19 & 2.86 & 2.86 \\
\hline
\end{tabular}
    \caption{Time (mins) \& memory (MB) required on different UEA MTSC datasets to achieve the best accuracy.}
    \label{tab:time_aftercs}
\end{table}\end{landscape}

\section{Minority Class Classification Report for Synthetic Datasets}

Classification report for minority class study in synthetic datasets having 100 channels and 100 signal length, with varying level of class imbalance.

\subsection{Moving Signal: No channel Selection}

This section presents the performance of standalone ROCKET on the minority class where the signal is moving randomly.

\subsubsection{Data Imbalance Ratio 90:10}

The below classification reports presents ROCKET performance on dataset with class imbalance of 90:10.

\begin{table}[H]
    \centering
\begin{tabular}{ccccc}
\hline
& precision & recall & f1-score & support \\
\hline
0 & 0.52 & 1 & 0.68 & 52 \\
\bigskip
1 & 0 & 0 & 0 & 48 \\

accuracy & 0.52 & 0.52 & 0.52 & 0.52 \\
macro avg & 0.26 & 0.5 & 0.34 & 100 \\
weighted avg & 0.27 & 0.52 & 0.35 & 100 \\
\hline
\end{tabular}    
\caption{Classification report for moving signal in GP dataset with training dataset in 90:10 ratio.}
\label{tab:cr_gp_ncs_0.1}
\end{table}

\begin{table}[H]
    \centering
\begin{tabular}{ccccc}
\hline
& precision & recall & f1-score & support \\
\hline
0 & 0.50 & 1 & 0.67 & 50.00 \\
\bigskip
1 & 0.00 & 0.00 & 0.00 & 50.00 \\
accuracy & 0.50 & 0.50 & 0.50 & 0.50 \\
macro avg & 0.25 & 0.50 & 0.33 & 100.00 \\
weighted avg & 0.25 & 0.50 & 0.33 & 100.00 \\
\hline
\end{tabular}    \caption{Classification report for moving signal in PP dataset with training dataset in 90:10 ratio.}
    \label{tab:cr_pp_ncs_0.1}
\end{table}

\begin{table}[H]
    \centering
\begin{tabular}{ccccc}
\hline
& precision & recall & f1-score & support \\
\hline
0 & 0.41 & 1.00 & 0.58 & 41.00 \\
\bigskip
1 & 0.00 & 0.00 & 0.00 & 59.00 \\
accuracy & 0.41 & 0.41 & 0.41 & 0.41 \\
macro avg & 0.21 & 0.50 & 0.29 & 100.00 \\
weighted avg & 0.17 & 0.41 & 0.24 & 100.00 \\
\hline
\end{tabular}
    \caption{Classification report for moving signal in AR dataset with training dataset in 90:10 ratio.}
    \label{tab:cr_pp_ncs_0.1}
\end{table}

\subsubsection{Data Imbalance Ratio 80:20}

The below classification reports presents ROCKET performance on dataset with class imbalance of 80:20.

\begin{table}[H]
    \centering
\begin{tabular}{ccccc}
\hline
& precision & recall & f1-score & support \\
\hline
0 & 0.52 & 1.00 & 0.68 & 52.00 \\
\bigskip
1 & 0.00 & 0.00 & 0 & 48.00 \\
accuracy & 0.52 & 0.52 & 0.52 & 0.52 \\
macro avg & 0.26 & 0.50 & 0.34 & 100.00 \\
weighted avg & 0.27 & 0.52 & 0.36 & 100.00 \\
\hline
\end{tabular}
    \caption{Classification report for moving signal in GP dataset with training dataset in 80:20 ratio.}
    \label{tab:cr_gp_ncs_0.2}
\end{table}

\begin{table}[H]
    \centering
\begin{tabular}{ccccc}
\hline
& precision & recall & f1-score & support \\
\hline
0 & 0.51 & 1.00 & 0.67 & 50.00 \\
\bigskip
1 & 1.00 & 0.02 & 0.04 & 50.00 \\
accuracy & 0.51 & 0.51 & 0.51 & 0.51 \\
macro avg & 0.75 & 0.51 & 0.36 & 100.00 \\
weighted avg & 0.75 & 0.51 & 0.36 & 100.00 \\
\hline
\end{tabular}    
\caption{Classification report for moving signal in PP dataset with training dataset in 80:20 ratio.}
    \label{tab:cr_pp_ncs_0.2}
\end{table}

\begin{table}[H]
    \centering
\begin{tabular}{ccccc}
\hline
& precision & recall & f1-score & support \\
\hline
0 & 0.41 & 1.00 & 0.58 & 41.00 \\
\bigskip
1 & 0.00 & 0.00 & 0.00 & 59.00 \\
accuracy & 0.41 & 0.41 & 0.41 & 0.41 \\
macro avg & 0.21 & 0.50 & 0.29 & 100.00 \\
weighted avg & 0.17 & 0.41 & 0.24 & 100.00 \\
\hline
\end{tabular}
    \caption{Classification report for moving signal in PP dataset with training dataset in 80:20 ratio.}
    \label{tab:cr_ar_ncs_0.2}
\end{table}

\subsection{Moving Signal: Channel Selection}

This section presents the performance of  ROCKET with channel selection on minority class where signal is moving randomly.

\subsubsection{Data Imbalance Ratio 90:10}
\begin{table}[H]
    \centering
\begin{tabular}{ccccc}
\hline
 & precision & recall & f1-score & support \\
\hline
0 & 0.52 & 1.00 & 0.68 & 52.00 \\
\bigskip
1 & 0.00 & 0.00 & 0.00 & 48.00 \\
accuracy & 0.52 & 0.52 & 0.52 & 0.52 \\
macro avg & 0.26 & 0.50 & 0.34 & 100.00 \\
weighted avg & 0.27 & 0.52 & 0.36 & 100.00 \\
\hline
\end{tabular}
    \caption{Classification report for moving signal in GP dataset with training dataset in 90:10 ratio.}
    \label{tab:cr_gp_cs_0.1}
\end{table}

\begin{table}[H]
    \centering
\begin{tabular}{ccccc}
\hline
& precision & recall & f1-score & support \\
\hline
0 & 0.51 & 1.00 & 0.67 & 50.00 \\
\bigskip
1 & 1.00 & 0.02 & 0.04 & 50.00 \\
accuracy & 0.51 & 0.51 & 0.51 & 0.51 \\
macro avg & 0.75 & 0.51 & 0.36 & 100.00 \\
weighted avg & 0.75 & 0.51 & 0.36 & 100.00 \\
\hline
\end{tabular}
    \caption{Classification report for moving signal in PP dataset with training dataset in 90:10 ratio.}
    \label{tab:cr_pp_cs_0.1}
\end{table}

\begin{table}[H]
    \centering
\begin{tabular}{ccccc}
\hline
& precision & recall & f1-score & support \\
\hline
0 & 0.41 & 1.00 & 0.58 & 41.00 \\
\bigskip
1 & 0.00 & 0.00 & 0.00 & 59.00 \\
accuracy & 0.41 & 0.41 & 0.41 & 0.41 \\
macro avg & 0.21 & 0.50 & 0.29 & 100.00 \\
weighted avg & 0.17 & 0.41 & 0.24 & 100.00 \\
\hline
\end{tabular}
    \caption{Classification report for moving signal in PP dataset with training dataset in 90:10 ratio.}
    \label{tab:cr_ar_cs_0.1}
\end{table}

\subsubsection{Data imbalance ratio 80:20}

\begin{table}[H]
    \centering
\begin{tabular}{ccccc}
\hline
& precision & recall & f1-score & support \\
\hline
0 & 0.54 & 1.00 & 0.70 & 52.00 \\
\bigskip
1 & 1.00 & 0.06 & 0.12 & 48.00 \\
accuracy & 0.55 & 0.55 & 0.55 & 0.55 \\
macro avg & 0.77 & 0.53 & 0.41 & 100.00 \\
weighted avg & 0.76 & 0.55 & 0.42 & 100.00 \\
\hline
\end{tabular}
    \caption{Classification report for moving signal in GP dataset with training dataset in 90:10 ratio.}
    \label{tab:cr_gp_cs_0.2}
\end{table}

\begin{table}[H]
    \centering
\begin{tabular}{ccccc}
\hline
& precision & recall & f1-score & support \\
\hline
0 & 0.60 & 1.00 & 0.75 & 50.00 \\
\bigskip
1 & 1.00 & 0.32 & 0.48 & 50.00 \\
accuracy & 0.66 & 0.66 & 0.66 & 0.66 \\
macro avg & 0.80 & 0.66 & 0.62 & 100.00 \\
weighted avg & 0.80 & 0.66 & 0.62 & 100.00 \\
\hline
\end{tabular}    
\caption{Classification report for moving signal in PP dataset with training dataset in 80:20 ratio.}
    \label{tab:cr_pp_cs_0.2}
\end{table}

\begin{table}[H]
    \centering
\begin{tabular}{ccccc}
\hline
& precision & recall & f1-score & support \\
\hline
0 & 0.42 & 1.00 & 0.59 & 41.00 \\
\bigskip
1 & 1.00 & 0.05 & 0.10 & 59.00 \\
accuracy & 0.44 & 0.44 & 0.44 & 0.44 \\
macro avg & 0.71 & 0.53 & 0.35 & 100.00 \\
weighted avg & 0.76 & 0.44 & 0.30 & 100.00 \\
\hline
\end{tabular}
    \caption{Classification report for moving signal in AR dataset with training dataset in 80:20 ratio.}
    \label{tab:cr_ar_cs_0.2}
\end{table}

\subsection{Stationary Signal: No channel Selection}
This section presents the performance of standalone ROCKET on minority class where signal stationary.

\subsubsection{Data Imbalance Ratio 90:10}

The below classification reports presents ROCKET performance on dataset with class imbalance of 90:10.

\begin{table}[H]
    \centering
\begin{tabular}{ccccc}
\hline
& precision & recall & f1-score & support \\
\hline
0 & 0.74 & 1.00 & 0.85 & 52.00 \\
\bigskip
1 & 1.00 & 0.63 & 0.77 & 48.00 \\
accuracy & 0.82 & 0.82 & 0.82 & 0.82 \\
macro avg & 0.87 & 0.81 & 0.81 & 100.00 \\
weighted avg & 0.87 & 0.82 & 0.81 & 100.00 \\
\hline
\end{tabular}
    \caption{Classification report for moving signal in GP dataset with training dataset in
90:10 ratio}
    \label{tab:cr_gp_ncss_0.1}
\end{table}

\begin{table}[H]
    \centering
\begin{tabular}{ccccc}
\hline
& precision & recall & f1-score & support \\
\hline
0 & 1.00 & 1.00 & 1.00 & 53.00 \\
\bigskip
1 & 1.00 & 1.00 & 1.00 & 47.00 \\
accuracy & 1.00 & 1.00 & 1.00 & 1.00 \\
macro avg & 1.00 & 1.00 & 1.00 & 100.00 \\
weighted avg & 1.00 & 1.00 & 1.00 & 100.00 \\
\hline
\end{tabular}
    \caption{Classification report for moving signal in PP dataset with training dataset in
90:10 ratio}
    \label{tab:cr_pp_ncss_0.1}
\end{table}

\begin{table}[H]
    \centering
\begin{tabular}{ccccc}
\hline
& precision & recall & f1-score & support \\
\hline
0 & 0.70 & 1.00 & 0.82 & 51.00 \\
\bigskip
1 & 1.00 & 0.55 & 0.71 & 49.00 \\
accuracy & 0.78 & 0.78 & 0.78 & 0.78 \\
macro avg & 0.85 & 0.78 & 0.77 & 100.00 \\
weighted avg & 0.85 & 0.78 & 0.77 & 100.00 \\
\hline
\end{tabular}
    \caption{Classification report for moving signal in AR dataset with training dataset in
90:10 ratio}
    \label{tab:cr_ar_ncss_0.1}
\end{table}

\subsubsection{Data Imbalance Ratio 80:20}

The below classification reports presents ROCKET performance on dataset with class imbalance of 80:20.

\begin{table}[H]
    \centering
\begin{tabular}{ccccc}
\hline
& precision & recall & f1-score & support \\
\hline
0 & 0.90 & 1.00 & 0.95 & 52.00 \\
\bigskip
1 & 1.00 & 0.88 & 0.93 & 48.00 \\
accuracy & 0.94 & 0.94 & 0.94 & 0.94 \\
macro avg & 0.95 & 0.94 & 0.94 & 100.00 \\
weighted avg & 0.95 & 0.94 & 0.94 & 100.00 \\
\hline
\end{tabular}
    \caption{Classification report for stationary signal in GP dataset with training dataset in 80:20 ratio.}
    \label{tab:cr_gp_ncss_0.2}
\end{table}

\begin{table}[H]
    \centering
\begin{tabular}{ccccc}
\hline
& precision & recall & f1-score & support \\
\hline
0 & 1.00 & 1.00 & 1.00 & 53.00 \\
\bigskip
1 & 1.00 & 1.00 & 1.00 & 47.00 \\
accuracy & 1.00 & 1.00 & 1.00 & 1.00 \\
macro avg & 1.00 & 1.00 & 1.00 & 100.00 \\
weighted avg & 1.00 & 1.00 & 1.00 & 100.00 \\
\hline
\end{tabular}
    \caption{Classification report for stationary signal in PP dataset with training dataset in 80:20 ratio.}
    \label{tab:cr_pp_ncss_0.2}
\end{table}

\begin{table}[H]
    \centering
\begin{tabular}{ccccc}
\hline
& precision & recall & f1-score & support \\
\hline
0.00 & 0.96 & 1.00 & 0.98 & 51.00 \\
\bigskip
1.00 & 1.00 & 0.96 & 0.98 & 49.00 \\
accuracy & 0.98 & 0.98 & 0.98 & 0.98 \\
macro avg & 0.98 & 0.98 & 0.98 & 100.00 \\
weighted avg & 0.98 & 0.98 & 0.98 & 100.00 \\
\hline
\end{tabular}
    \caption{Classification report for stationary signal in AR dataset with training dataset in 80:20 ratio.}
    \label{tab:cr_ar_ncss_0.2}
\end{table}

\subsection{Stationary Signal: Channel Selection}
This section presents the performance of  ROCKET with channel selection on minority class where signal is stationary.

\subsubsection{Data Imbalance Ratio 90:10}
The below classification reports presents ROCKET performance on dataset with class imbalance of 90:10.

\begin{table}[H]
    \centering
\begin{tabular}{ccccc}
\hline
& precision & recall & f1-score & support \\
\hline
0 & 0.96 & 1.00 & 0.98 & 52.00 \\
\bigskip
1 & 1.00 & 0.96 & 0.98 & 48.00 \\
accuracy & 0.98 & 0.98 & 0.98 & 0.98 \\
macro avg & 0.98 & 0.98 & 0.98 & 100.00 \\
weighted avg & 0.98 & 0.98 & 0.98 & 100.00 \\
\hline
\end{tabular}
    \caption{Classification report for stationary signal in GP dataset with training dataset in 90:10 ratio.}
    \label{tab:cr_gp_css_0.1}
\end{table}

\begin{table}[H]
    \centering
    \begin{tabular}{ccccc}
\hline
& precision & recall & f1-score & support \\
\hline
0 & 1.00 & 1.00 & 1.00 & 53.00 \\
\bigskip
1 & 1.00 & 1.00 & 1.00 & 47.00 \\
accuracy & 1.00 & 1.00 & 1.00 & 1.00 \\
macro avg & 1.00 & 1.00 & 1.00 & 100.00 \\
weighted avg & 1.00 & 1.00 & 1.00 & 100.00 \\
\hline
\end{tabular}
    \caption{Classification report for stationary signal in PP dataset with training dataset in 90:10 ratio.}
    \label{tab:cr_pp_css_0.1}
\end{table}

\begin{table}[H]
    \centering
\begin{tabular}{ccccc}
\hline
& precision & recall & f1-score & support \\
\hline
0 & 0.91 & 1.00 & 0.95 & 51.00 \\
\bigskip
1 & 1.00 & 0.90 & 0.95 & 49.00 \\
accuracy & 0.95 & 0.95 & 0.95 & 0.95 \\
macro avg & 0.96 & 0.95 & 0.95 & 100.00 \\
weighted avg & 0.95 & 0.95 & 0.95 & 100.00 \\
\hline
\end{tabular}
    \caption{Classification report for stationary signal in AR dataset with training dataset in 90:10 ratio.}
    \label{tab:cr_ar_css_0.1}
\end{table}

\subsubsection{Data Imbalance Ratio 80:20}
The below classification reports presents ROCKET performance on dataset with class imbalance of 80:20.
\begin{table}[H]
    \centering
\begin{tabular}{ccccc}
\hline
& precision & recall & f1-score & support \\
\hline
0 & 1.00 & 1.00 & 1.00 & 52.00 \\
\bigskip
1 & 1.00 & 1.00 & 1.00 & 48.00 \\
accuracy & 1.00 & 1.00 & 1.00 & 1.00 \\
macro avg & 1.00 & 1.00 & 1.00 & 100.00 \\
weighted avg & 1.00 & 1.00 & 1.00 & 100.00 \\
\hline
\end{tabular}
    \caption{Classification report for stationary signal in GP dataset with training dataset in 80:20 ratio.}
    \label{tab:my_label}
\end{table}

\begin{table}[H]
    \centering
\begin{tabular}{ccccc}
\hline
& precision & recall & f1-score & support \\
\hline
0 & 1.00 & 1.00 & 1.00 & 53.00 \\
\bigskip
1 & 1.00 & 1.00 & 1.00 & 47.00 \\
accuracy & 1.00 & 1.00 & 1.00 & 1.00 \\
macro avg & 1.00 & 1.00 & 1.00 & 100.00 \\
weighted avg & 1.00 & 1.00 & 1.00 & 100.00 \\
\hline
\end{tabular}
    \caption{Classification report for stationary signal in PP dataset with training dataset in 80:20 ratio.}
    \label{tab:my_label}
\end{table}

\begin{table}[H]
    \centering
\begin{tabular}{ccccc}
\hline
& precision & recall & f1-score & support \\
\hline
0 & 0.98 & 1.00 & 0.99 & 51.00 \\
\bigskip
1 & 1.00 & 0.98 & 0.99 & 49.00 \\
accuracy & 0.99 & 0.99 & 0.99 & 0.99 \\
macro avg & 0.99 & 0.99 & 0.99 & 100.00 \\
weighted avg & 0.99 & 0.99 & 0.99 & 100.00 \\
\hline
\end{tabular}
    \caption{Classification report for stationary signal in AR dataset with training dataset in 80:20 ratio.}
    \label{tab:my_label}
\end{table}

\begin{table}[H]
    \centering
\begin{tabular}{ccccccc}
\hline
Channel Selection $\rightarrow $ & \multicolumn{3}{c}{ECS} &  \multicolumn{3}{c}{ECP} \\
\hline
Class Prototype $\rightarrow$ & Mean & Median & MAD &  Mean & Median & MAD \\
\hline
Distance function $\rightarrow$& EU $\vert$ DTW & EU $\vert$ DTW & EU $\vert$ DTW &  EU $\vert$ DTW & EU $\vert$ DTW & EU $\vert$ DTW \\
\hline
Rocket & 67.17 $\vert$ 67.17 & 66.56 $\vert$ 66.56 & 67.05 $\vert$ 67.05 & 71.58 $\vert$ 71.53 & 71.69 $\vert$ 71.60 &  71.86 $\vert$ 71.58 \\
WeaselMuse & 66.48 $\vert$ 68.18 &  64.1 $\vert$ 64.1 &  64.51 $\vert$ 64.51 & 68.71 $\vert$ 70.30  & 68.96 $\vert$ 68.56 & 68.94 $\vert$ 68.62\\
MrSEQL-SAX & 63.14 $\vert$ 63.14  & 63.32 $\vert$ 63.32 & 63.68 $\vert$ 63.68 & 67.44 $\vert$ 66.63  & 67.35 $\vert$ 66.83 & 67.37 $\vert$ 66.79 \\
\hline
\end{tabular}
    \caption{Accuracy for channel selection using Euclidean vs DTW distance.}
    \label{tab:dtweu_acc}
\end{table}

\begin{table}[H]
    \centering
\begin{tabular}{cccc}
\hline
Channel Selection $\rightarrow $ &  \multicolumn{3}{c}{ECP} \\
\hline
Class Prototype $\rightarrow$ &  Mean & Median & MAD \\
\hline
Distance function $\rightarrow$ &   EU $\vert$ DTW & EU $\vert$ DTW & EU $\vert$ DTW \\
\hline
Rocket & 23.13 $\vert$ 25 & 26.40 $\vert$ 28.24 & 26.74 $\vert$ 28.79  \\
WeaselMuse & 871.85  $\vert$ 988.55 &  843.87 $\vert$ 1027.60 &  634.50 $\vert$ 985.07\\
MrSEQL-SAX & 2067.11 $\vert$ 2004.19  & 2117.61 $\vert$ 2358.81 & 1178.24 $\vert$ 2269.91 \\
\hline
\end{tabular}
    \caption{Euclidean vs DTW distance computation time (minutes) (Part ECS)}
    \label{tab:dtweu_time}
\end{table}

\begin{table}[H]
    \centering
\begin{tabular}{cccc}
\hline
Channel Selection $\rightarrow$  &  \multicolumn{3}{c}{ECP} \\
\hline
Class Prototype $\rightarrow$ & Mean & Median & MAD \\
\hline
Distance function $\rightarrow$ & EU $\vert$ DTW & EU $\vert$ DTW & EU $\vert$ DTW \\
\hline
Rocket & 31.69 $\vert$ 52.02 & 30.17 $\vert$ 49.63 &  30.4 $\vert$ 48.38 \\
WeaselMuse  & 1208.8 $\vert$ 1700.2  & 1206.94 $\vert$ 1294.82 & 966.51 $\vert$ 1271.76\\
MrSEQL-SAX &  2951.17 $\vert$ 3684.17  & 2950.34 $\vert$ 3262.07 & 1674.18 $\vert$ 3303.49 \\
\hline
\end{tabular}
    \caption{EU vs DTW computation time (minutes) (Part ECP)}
    \label{tab:dtweu_time}
\end{table}

\begin{figure}[H]
    \centering
    \includegraphics[width=\textwidth]{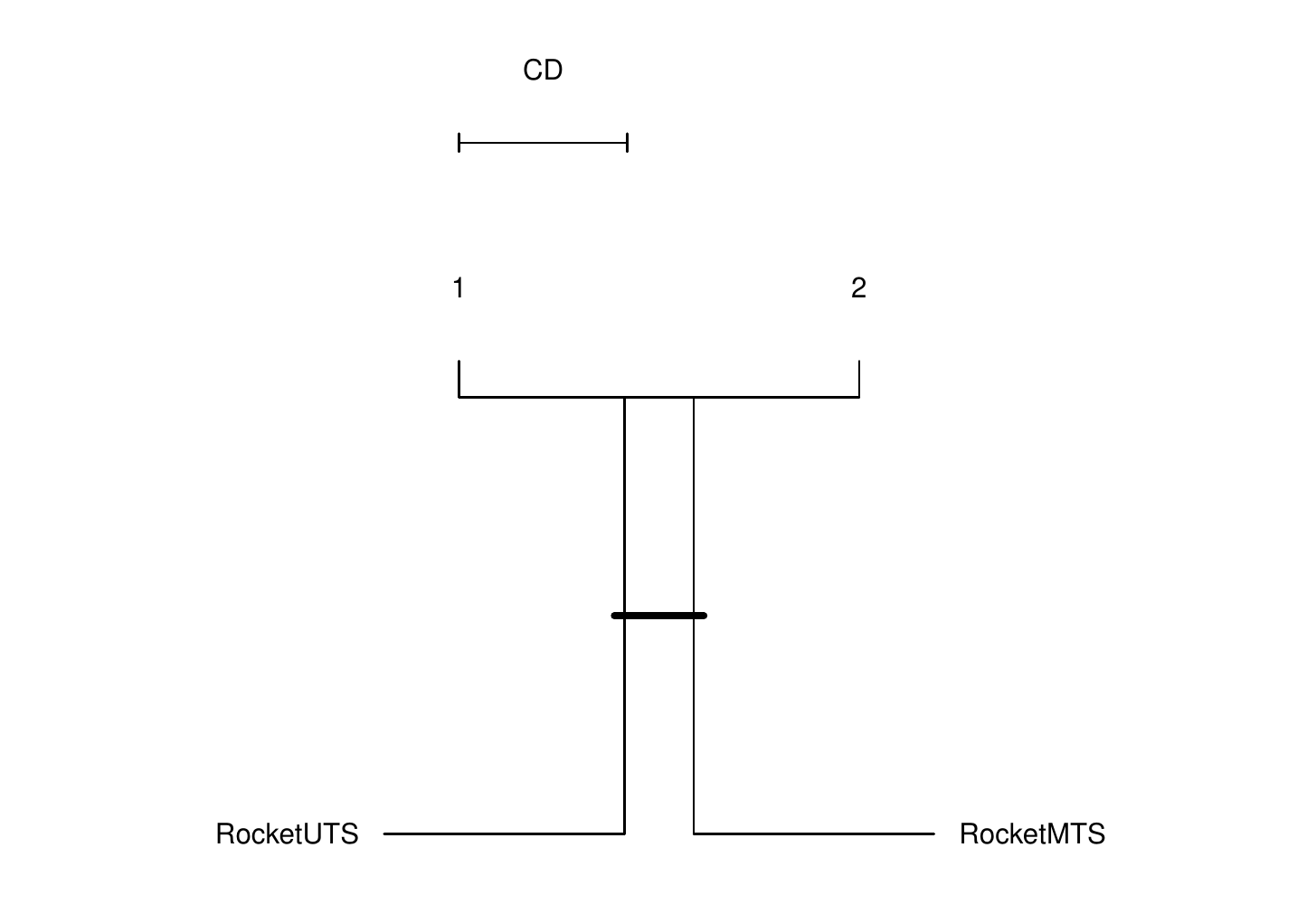}
    \caption{Examination of channel interdependence on 23 UEA datasets. There is no significant difference in accuracy between using kernels across all channels versus single-channel kernels.}
    \label{fig:utsvsmtsrockt}
\end{figure}

\begin{figure}[H]
    \centering
    \includegraphics[width=\textwidth]{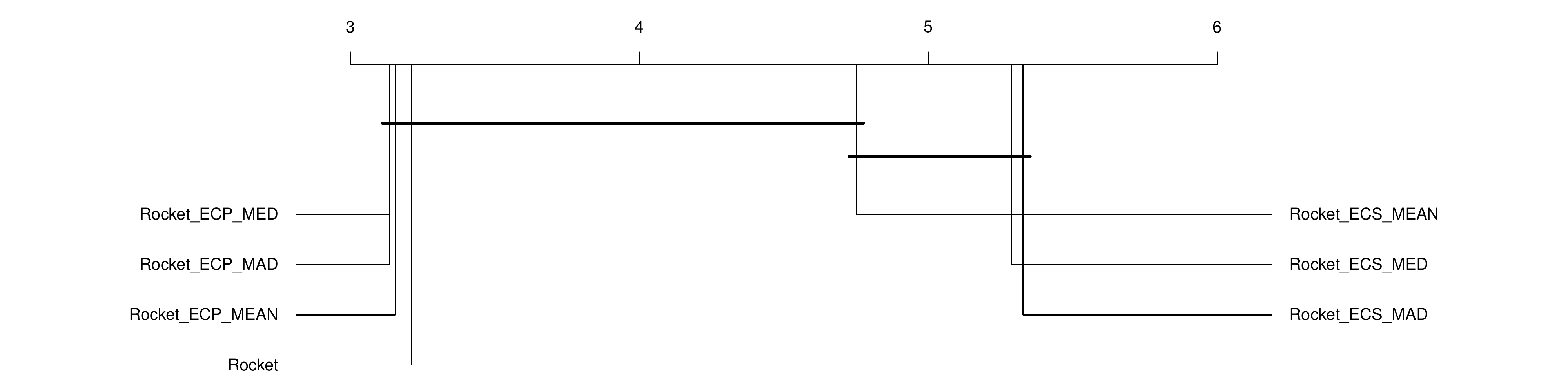}
    \caption{Critical difference diagram of Rocket with and without channel selection.}
    \label{fig:rocket_cd}
\end{figure}

\begin{figure}[H]
    \centering
    \includegraphics[width=\textwidth]{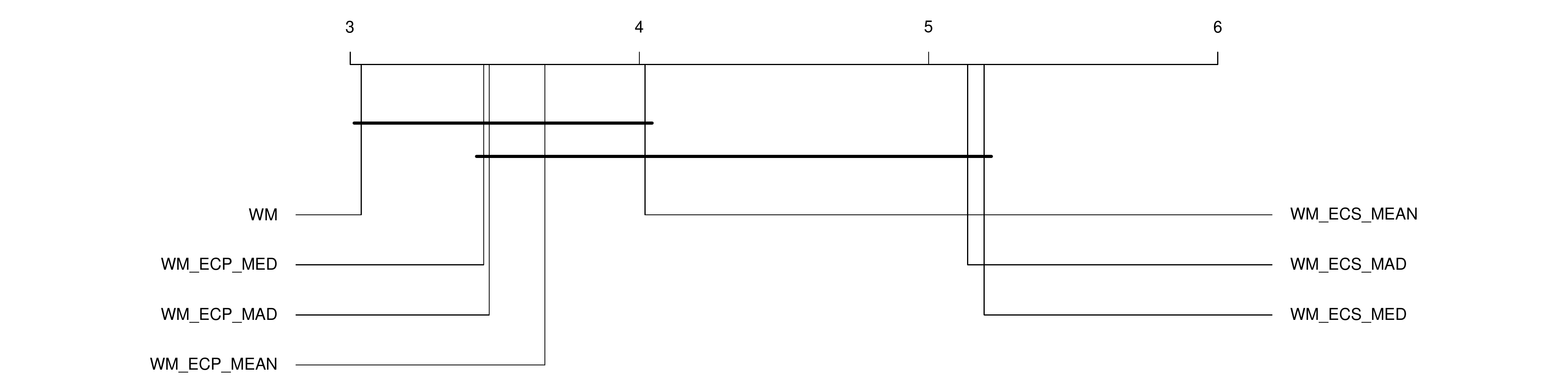}
    \caption{Critical difference diagram of Weasel Muse with and without channel selection.}
    \label{fig:wm_cd}
\end{figure}

\begin{figure}[H]
    \centering
    \includegraphics[width=\textwidth]{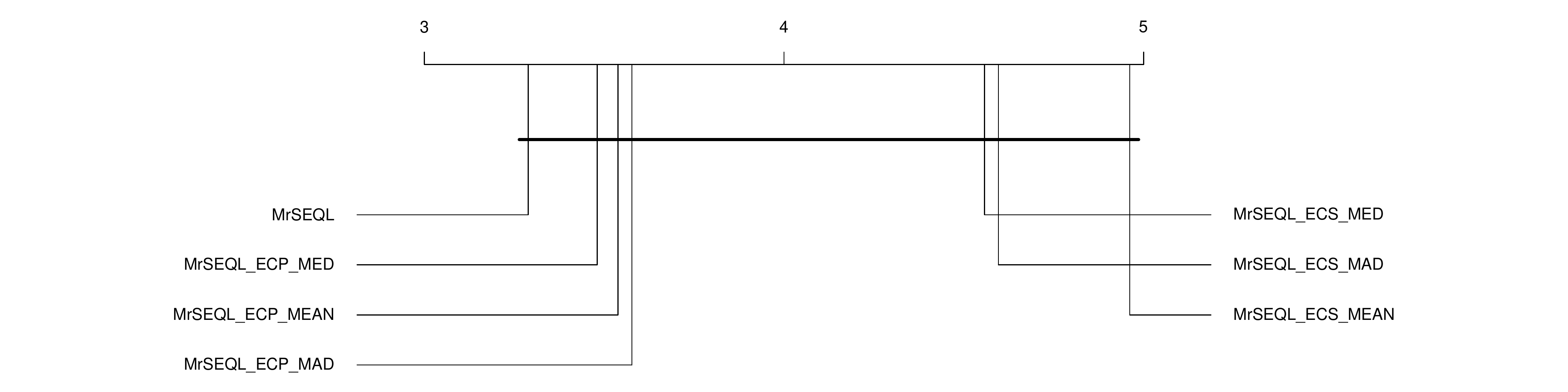}
    \caption{Critical difference diagram of MrSEQL with and without channel selection.}
    \label{fig:mrseql_cd}
\end{figure}

\begin{table}[H]
    \centering
\begin{tabular}{|c|c|}
\hline
dataset & class counts \\
\hline
ArticularyWordRecognition & 
    \makecell[l]{\{'1.0': 12, '2.0': 12, '3.0': 12, '4.0': 12, '5.0': 12, '6.0': 12, '7.0': 12, \\'8.0': 12, '9.0': 12, '10.0': 12, '11.0': 12, '12.0': 12, '13.0': 12, '14.0': 12, \\'15.0': 12, '16.0': 12, '17.0': 12, '18.0': 12, '19.0': 12, '20.0': 12, '21.0': 12,\\ '22.0': 12, '23.0': 12, '24.0': 12, '25.0': 12\}} \\
\hline
AtrialFibrillation & \makecell[l]{\{'n': 5, 's': 5, 't': 5\}} \\
\hline
BasicMotions & \makecell[l]{\{'standing': 10, 'running': 10, 'walking': 10, 'badminton': 10\}} \\
\hline
Cricket & \makecell[l]{\{'1.0': 6, '2.0': 6, '3.0': 6, '4.0': 6, '5.0': 6, \\'6.0': 6, '7.0': 6, '8.0': 6, '9.0': 6, '10.0': 6, '11.0': 6, '12.0': 6\}} \\
\hline
DuckDuckGeese & \makecell[l]{\{'black\-bellied\_whistling\_duck': 10, 'canadian\_goose': 10,\\ 'greylag\_goose': 10, 'pink\-footed\_goose': 10,\\ 'white\-faced\_whistling\_duck': 10\}} \\
\hline
ERing & \makecell[l]{\{'2': 45, '5': 45, '3': 45, '6': 45, '1': 45, '4': 45\}} \\
\hline
EigenWorms & \makecell[l]{\{'1': 55, '2': 22, '3': 18, '4': 23, '5': 13\}} \\
\hline
Epilepsy & \makecell[l]{\{'epilepsy': 34, 'walking': 37, 'running': 37, 'sawing': 30\}} \\
\hline
EthanolConcentration & \makecell[l]{\{'e35': 66, 'e38': 66, 'e40': 66, 'e45': 65\}} \\
\hline
FaceDetection & \makecell[l]{\{'0': 1762, '1': 1762\}} \\
\hline
FingerMovements & \makecell[l]{\{'right': 51, 'left': 49\}} \\
\hline
HandMovementDirection & \makecell[l]{\{'right': 14, 'forward': 30, 'left': 15, 'backward': 15\}} \\
\hline
Handwriting & \makecell[l]{\{'15.0': 29, '1.0': 36, '23.0': 32, '9.0': 27, '2.0': 34, '17.0': 32,\\ '14.0': 35, '24.0': 37, '25.0': 26, '11.0': 25, '12.0': 37, '7.0': 29, '3.0': 29, \\'21.0': 34, '6.0': 30, '20.0': 31, '5.0': 34, '4.0': 43, '26.0': 32, \\'8.0': 32, '16.0': 33, '13.0': 40, '19.0': 33, '10.0': 26, '18.0': 31, '22.0': 43\}} \\
\hline
Heartbeat & \makecell[l]{\{'normal': 57, 'abnormal': 148\}} \\
\hline
LSST & \makecell[l]{\{'6': 35, '15': 124, '16': 270, '42': 382, '52': 63, '53': 7, '62': 153, \\'64': 24, '65': 313, '67': 68, '88': 121, '90': 777, '92': 77, '95': 52\}} \\
\hline
Libras & \makecell[l]{\{'1': 12, '2': 12, '3': 12, '4': 12, '5': 12, '6': 12, \\'7': 12, '8': 12, '9': 12, '10': 12, '11': 12, '12': 12, '13': 12, '14': 12, '15': 12\}} \\
\hline
MotorImagery & \makecell[l]{\{'tongue': 50, 'finger': 50\}} \\
\hline
NATOPS & \makecell[l]{\{'4.0': 30, '5.0': 30, '6.0': 30, '1.0': 30, '3.0': 30, '2.0': 30\}} \\
\hline
PEMS-SF & \makecell[l]{\{'4.0': 23, '2.0': 25, '7.0': 20, '3.0': 26, '1.0': 30, '5.0': 22, '6.0': 27\}} \\
\hline
PenDigits & \makecell[l]{\{'8': 336, '9': 336, '1': 364, '4': 364, '7': 364, \\'0': 363, '2': 364, '5': 335, '3': 336, '6': 336\}} \\
\hline
PhonemeSpectra & \makecell[l]{\{'aa': 86, 'ae': 86, 'ah': 86, 'ao': 86, 'aw': 86, 'ay': 86, 'b': 86, 'ch': 86, \\'d': 86, 'dh': 86, 'eh': 86, 'er': 86, 'ey': 86, 'f': 86, 'g': 86, 'hh': 86, \\'ih': 86, 'iy': 86, 'jh': 86, 'k': 86, 'l': 86, 'm': 86, 'n': 86, 'ng': 86, 'ow': 86,\\ 'oy': 86, 'p': 86, 'r': 86, 's': 86, 'sh': 86, 't': 86, 'th': 86, 'uh': 86, 'uw': 86, \\'v': 86, 'w': 86, 'y': 86, 'z': 86, 'zh': 85\}} \\
\hline
RacketSports & \makecell[l]{\{'badminton\_smash': 40, 'badminton\_clear': 43, \\ 'squash\_forehandboast': 35, 'squash\_backhandboast': 34\}} \\
\hline
SelfRegulationSCP1 & \makecell[l]{\{'positivity': 146, 'negativity': 147\}} \\
\hline
SelfRegulationSCP2 & \makecell[l]{\{'positivity': 90, 'negativity': 90\}} \\
\hline
StandWalkJump & \makecell[l]{\{'standing': 5, 'walking': 5, 'jumping': 5\}} \\
\hline
UWaveGestureLibrary & \makecell[l]{\{'1.0': 40, '2.0': 40, '3.0': 40, '4.0': 40, '5.0': 40, '6.0': 40, '7.0': 40, '8.0': 40\}} \\
\hline
\end{tabular}
    \caption{Class counts for UEA dataset.}
    \label{tab:uea_cc}
\end{table}

\begin{figure}
\centering
\subfloat[]{\includegraphics[width=0.5\textwidth]{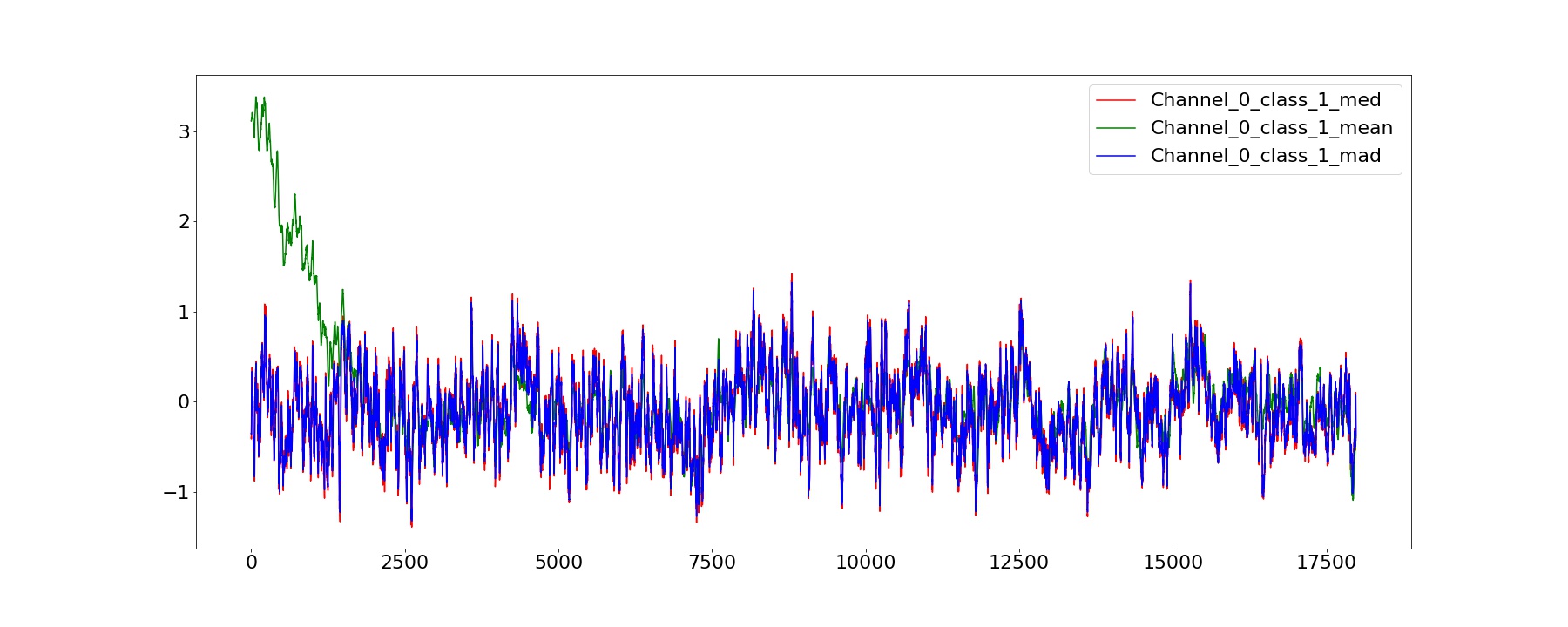}}
\subfloat[]{\includegraphics[width=0.5\textwidth]{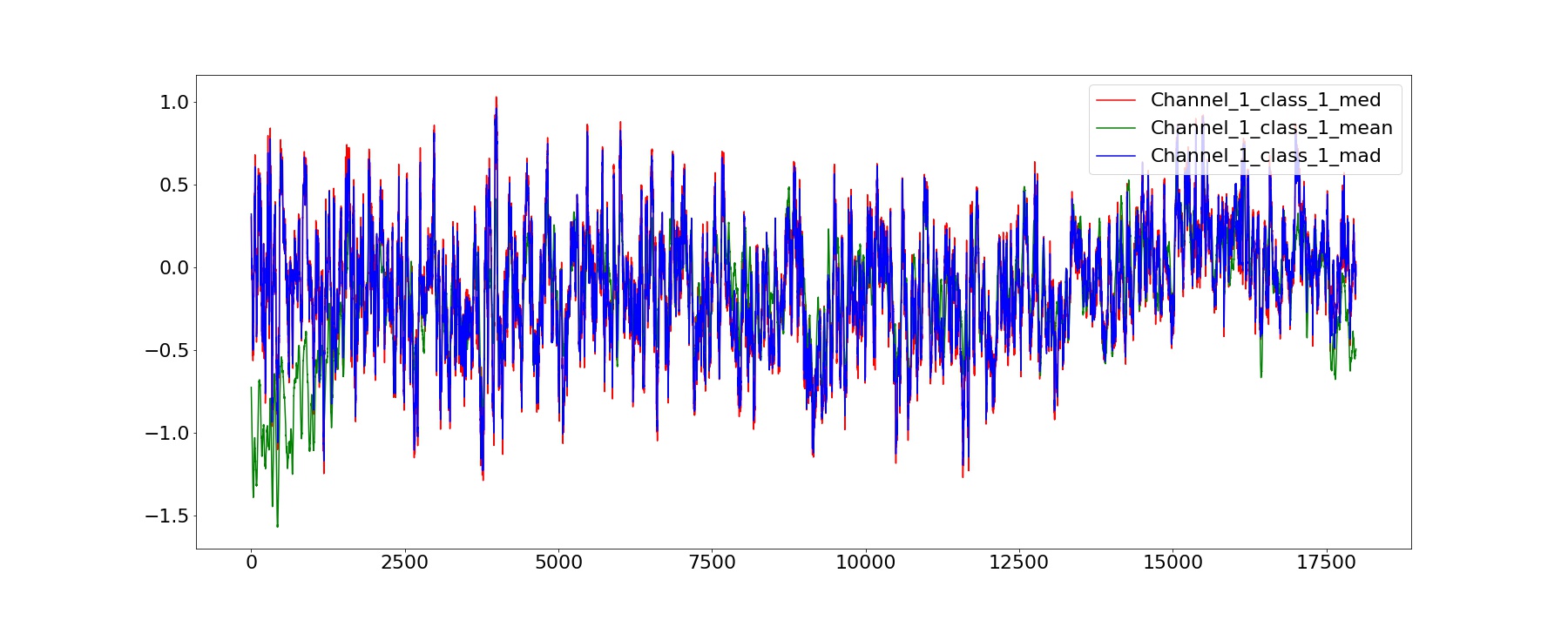}}
\hfill
\subfloat[]{\includegraphics[width=0.5\textwidth]{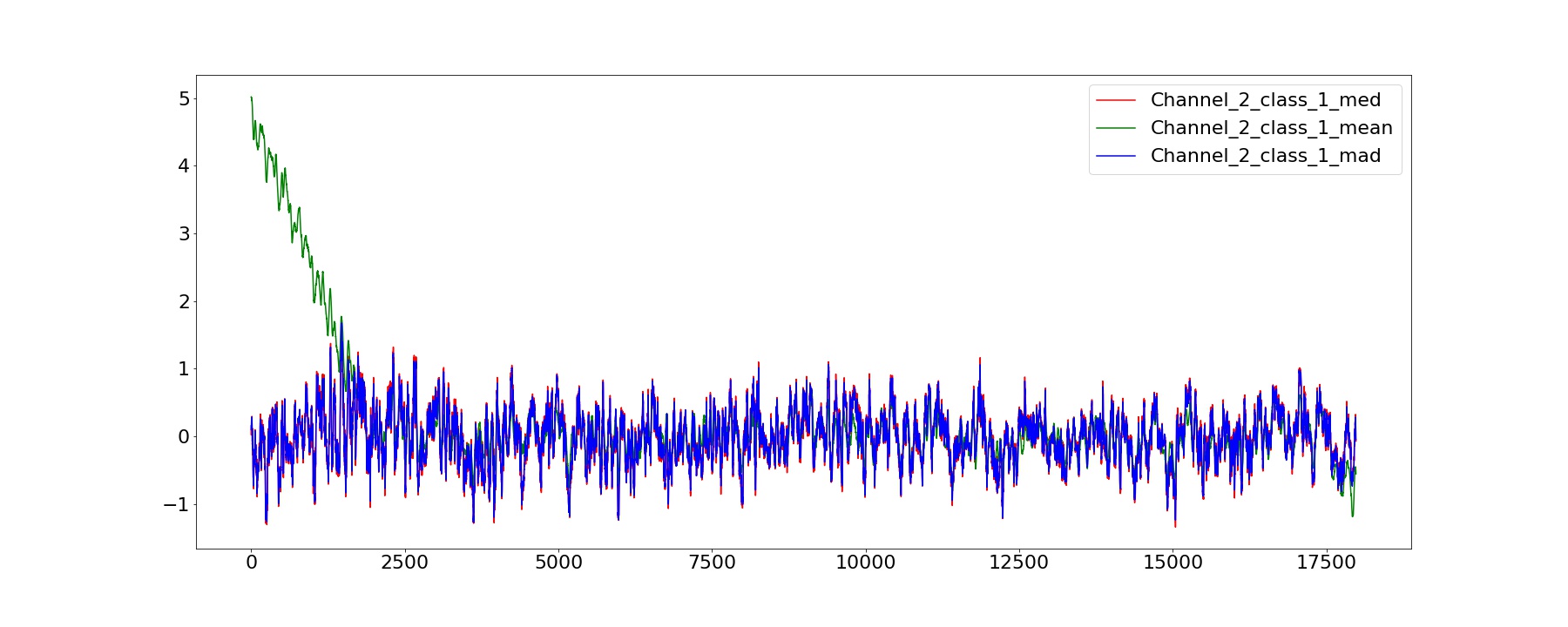}}
\subfloat[]{\includegraphics[width=0.5\textwidth]{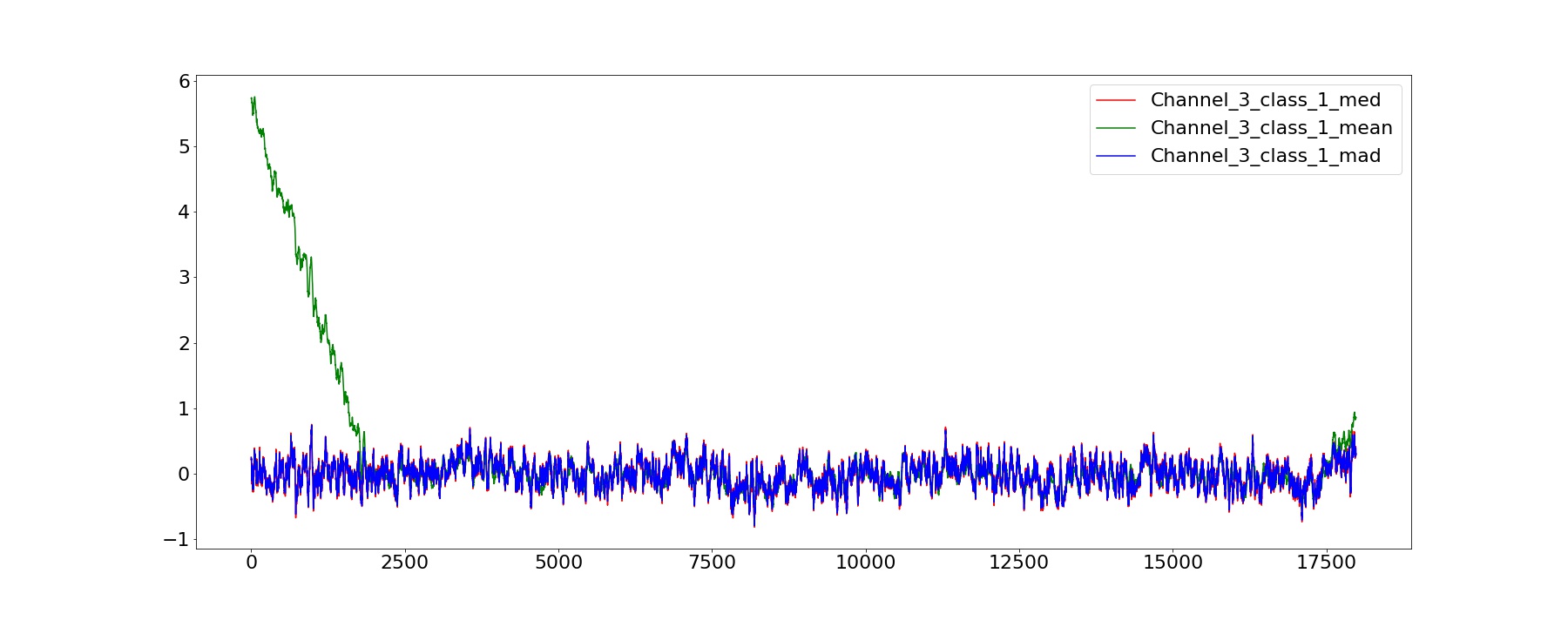}}
\hfill
\subfloat[]{\includegraphics[width=0.5\textwidth]{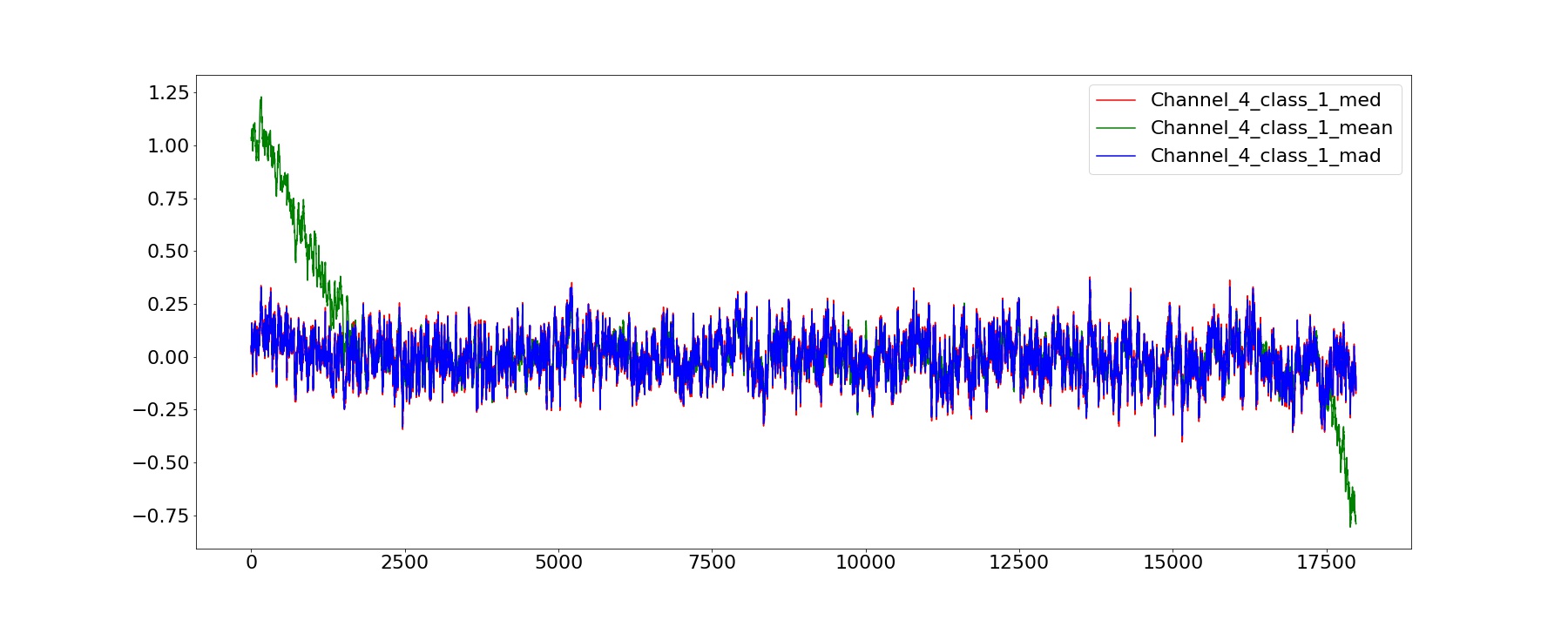}}
\subfloat[]{\includegraphics[width=0.5\textwidth]{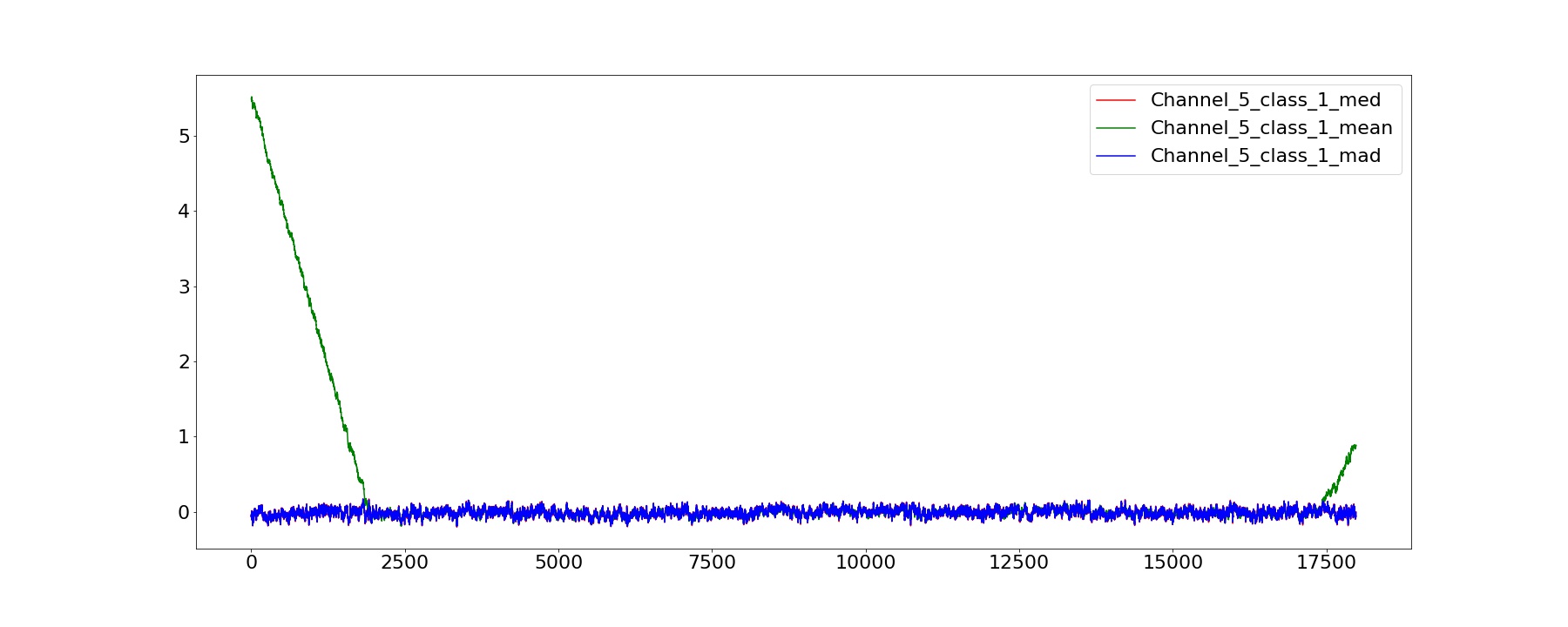}}

\caption[]{Class prototype for 6 channels in class 1 for EW dataset.}
\label{fig:cpew}
\end{figure}

\begin{table}[H]
    \centering
    \begin{tabular}{ccccc}
    \hline
    \multicolumn{1}{c}{Class Prototype$\rightarrow$}  &   \multicolumn{1}{c}{Mean} & \multicolumn{1}{c}{Median} & \multicolumn{1}{c}{MAD}\\
    \hline
    Classifiers$\rightarrow$ & Rocket $\lvert$ MrSEQL $\lvert$ WM  & Rocket $\lvert$ MrSEQL $\lvert$ WM & Rocket $\lvert$ MrSEQL $\lvert$ WM \\
    \hline
       ASR (3)    & \textcolor{blue}{+2.61}$\vert$ \textcolor{red}{-0.07}$\vert$ \textcolor{red}{-0.20} &  \textcolor{blue}{+3.38} $\vert$ \textcolor{blue}{+0.59} $\vert$ \textcolor{red}{-1.55} & \textcolor{blue}{+4.48} $\vert$ \textcolor{blue}{+1.26} $\vert$ \textcolor{red}{-1.55}\\
       ECG (2) & \textcolor{red}{-3.73} $\vert$ \textcolor{blue}{+10} $\vert$ \textcolor{red}{-16.66}
       &\textcolor{red}{-4.00} $\vert$ \textcolor{red}{-6.67} $\vert$ \textcolor{red}{-6.66} & \textcolor{red}{-4.00} $\vert$ \textcolor{red}{-6.67} $\vert$ \textcolor{red}{-6.66} \\
       EEG (6)  & \textcolor{blue}{+1.18} $\vert$ \textcolor{blue}{+0.45} $\vert$ \textcolor{red}{-0.64} & \textcolor{red}{-0.29} $\vert$ \textcolor{blue}{+1.15} $\vert$ \textcolor{red}{-2.12} & \textcolor{red}{-0.28} $\vert$ \textcolor{blue}{+1.10} $\vert$ \textcolor{red}{-2.29}\\
    HAR (9) & \textcolor{red}{-2.31} $\vert$ \textcolor{red}{-0.43} $\vert$ \textcolor{blue}{+0.11} &  \textcolor{red}{-1.20} $\vert$ \textcolor{red}{-0.55} $\vert$ \textcolor{blue}{+0.18} & \textcolor{red}{-1.20} $\vert$ \textcolor{red}{-0.55} $\vert$ \textcolor{blue}{+0.18}\\
    MC (3) & \textcolor{blue}{0.00}  $\vert$ \textcolor{red}{-0.48} $\vert$ \textcolor{blue}{+0.01} & \textcolor{blue}{0.00} $\vert$ \textcolor{red}{-0.24} $\vert$ \textcolor{blue}{+0.01}
    & \textcolor{blue}{0.00}  $\vert$ \textcolor{red}{-0.24} $\vert$ \textcolor{blue}{+0.01} \\
    Other (3) & \textcolor{blue}{+4.44} $\vert$ \textcolor{red}{-1.74} $\vert$ \textcolor{red}{-1.34} & \textcolor{blue}{+4.44} $\vert$ \textcolor{red}{-2.31} $\vert$ \textcolor{red}{-1.73}
          & \textcolor{blue}{+4.44} $\vert$ \textcolor{red}{-2.69} $\vert$ \textcolor{red}{-1.54}\\
         \hline
    \end{tabular}
    \caption{$\Delta$ Accuracy ECP channel selection for datasets from different domains.}
    \label{tab:domain}
\end{table}